%% file: main.tex
\PassOptionsToPackage{table,dvipsnames}{xcolor}
\documentclass[11pt,letterpaper,logo,onecolumn]{main}
\usepackage[numbers,sort&compress,square]{natbib}

\usepackage{microtype}

\usepackage{enumitem}
\usepackage{needspace}

\usepackage{amsmath,amssymb}

\usepackage{graphicx}
\usepackage{float}
\usepackage{placeins}

\usepackage{booktabs}
\usepackage{multirow}
\usepackage{array}
\usepackage{tabularx}
\usepackage{etoolbox}
\usepackage[normalem]{ulem}
\usepackage[most]{tcolorbox}
\definecolor{gblue9}{RGB}{23,78,166}
\usepackage{tikz}
\usepackage{listings}

\AtBeginEnvironment{table}{\setlength{\belowcaptionskip}{5pt}}
\AtBeginEnvironment{figure}{%
  \setlength{\abovecaptionskip}{6pt}%
  \setlength{\belowcaptionskip}{0pt}}

\definecolor{barfill}{HTML}{4E79A7}
\definecolor{barbg}{HTML}{ECECEC}
\definecolor{ratiolow}{HTML}{2E7355}
\definecolor{ratiohigh}{HTML}{B5532F}
\definecolor{caseframe}{HTML}{4E79A7}
\definecolor{casetitlebg}{HTML}{E8F0F9}
\definecolor{caselabel}{HTML}{2C5481}
\definecolor{caseask}{HTML}{B8860B}
\definecolor{caseanswer}{HTML}{117733}
\definecolor{caseslate}{HTML}{776480}

\newlength{\databarwidth}
\newlength{\databarhalfheight}
\newcommand{\databar}[3][barfill]{%
  {\footnotesize\begin{tikzpicture}[baseline=-0.5ex]
    \fill[barbg] (0,-\databarhalfheight) rectangle (\databarwidth,\databarhalfheight);
    \fill[#1] (0,-\databarhalfheight) rectangle ({#2/100*\databarwidth},\databarhalfheight);
  \end{tikzpicture}\hspace{0.35em}\makebox[4.2em][l]{#3}}%
}

\newcommand{\std}[1]{{\scriptsize$\,\pm#1$}}
\newcommand{\think}[1]{\,{\scriptsize\textcolor{black!55}{#1}}}
\newcommand{\modellogo}[1]{\raisebox{-0.25ex}{%
  \ifstrequal{#1}{gpt}{\includegraphics[height=1.9ex]{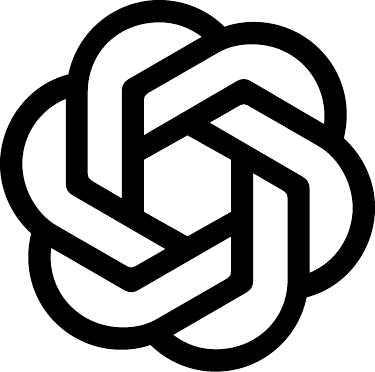}}{}%
  \ifstrequal{#1}{claude}{\includegraphics[height=1.9ex]{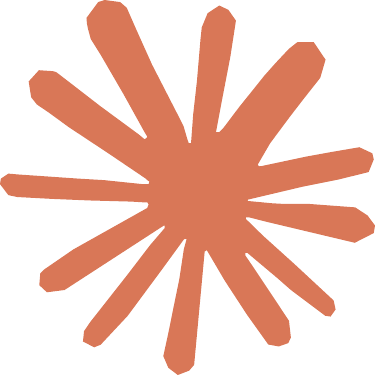}}{}%
  \ifstrequal{#1}{kimi}{\includegraphics[height=1.9ex]{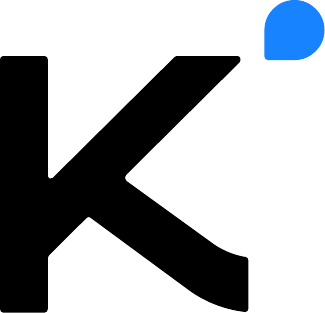}}{}%
  \ifstrequal{#1}{grok}{\includegraphics[height=1.9ex]{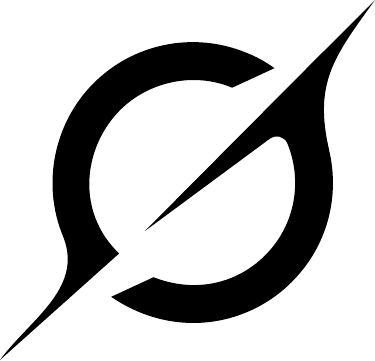}}{}%
  \ifstrequal{#1}{deepseek}{\includegraphics[height=1.9ex]{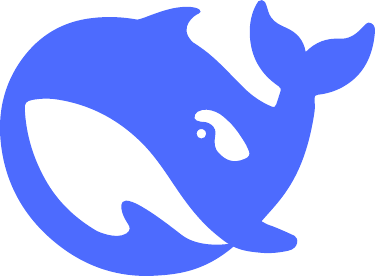}}{}%
  \ifstrequal{#1}{glm}{\includegraphics[height=1.9ex]{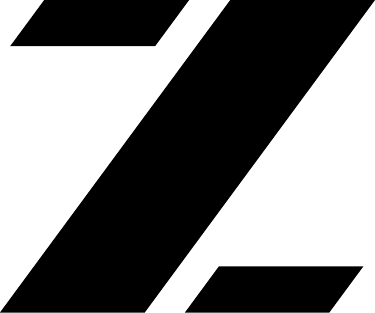}}{}%
  \ifstrequal{#1}{doubao}{\includegraphics[height=1.9ex]{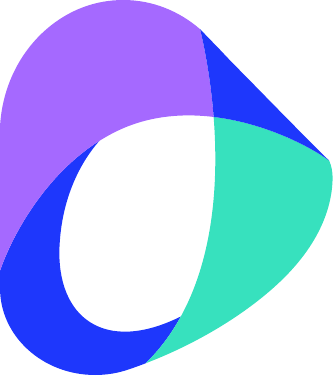}}{}}}
\newcommand{\withlogo}[2]{\makebox[1.2em][c]{\modellogo{#1}}\hspace{0.15em}#2}

\newcommand{\acc}[2]{%
  \makebox[2.65em][r]{$#1$}%
  \makebox[2.25em][l]{\if\relax\detokenize{#2}\relax\else\std{#2}\fi}%
}

\newcommand{\finding}[2]{%
  \par\Needspace{0.13\textheight}%
  \begin{tcolorbox}[
    enhanced,
    colback=black!3,
    colframe=black!28,
    boxrule=0.55pt,
    arc=0.8mm,
    left=5pt,
    right=6pt,
    top=4pt,
    bottom=4pt,
    before skip=8pt,
    after={\par\nobreak\vspace{5pt}},
    drop shadow=black!16]
    \begin{tabularx}{\linewidth}{@{}>{\raggedright\arraybackslash}p{4.6em}X@{}}
      {\small\bfseries Finding~#1} & \textbf{#2}
    \end{tabularx}
  \end{tcolorbox}\noindent\ignorespaces}

\fancypagestyle{titlefolio}[firststyle]{%
  \fancyfoot[L]{\footerfont
                The full author list appears at the end of the paper.\quad
                Correspondence:
                \texttt{mingzhang23@m.fudan.edu.cn},
                \texttt{maxmpan@tencent.com},
                \texttt{yangruizhi@fudan.edu.cn}, and
                \texttt{qz@fudan.edu.cn}.}%
  \fancyfoot[R]{\footerfont\thepage}}

\fancypagestyle{authorfolio}[fancy]{%
  \fancyfoot[L]{\footerfont\textsuperscript{*}Equal contribution.\quad
                \textsuperscript{\dag}Corresponding authors.}}

\hypersetup{colorlinks=true, linkcolor=black, citecolor=blue, urlcolor=blue}
\expandafter\def\expandafter\UrlBreaks\expandafter{\UrlBreaks\do\-}

\newcommand{\authorname}[1]{\textbf{#1}}

\title{HyperLogic: A Hard, Forward-Authored Chinese Logical Reasoning Benchmark
  with Execution-Derived Answers}

\author{
  \normalsize
  Fudan University
  \qquad
  Hunyuan Team, Tencent
}

\newcommand{\contributorlist}{%
  \authorname{Ming Zhang}\textsuperscript{*},
  \authorname{Qiyuan Peng}\textsuperscript{*},
  \authorname{Yinxi Wei},
  \authorname{Yujiong Shen},
  \authorname{Kexin Tan},
  \authorname{Yuhui Wang},
  \authorname{Zhenghao Xiang},
  \authorname{Junjie Ye},
  \authorname{Zhangyue Yin},
  \authorname{Zhiheng Xi},
  \authorname{Shihan Dou},
  \authorname{Weikang Wang},
  \authorname{Yuhao Zhang}}
\newcommand{\leadlist}{%
  \authorname{Tao Gui},
  \authorname{Ruizhi Yang}\textsuperscript{\dag},
  \authorname{Qi Zhang}\textsuperscript{\dag},
  \authorname{Xuanjing Huang},
  \authorname{Alex Chen},
  \authorname{Maxm Pan}\textsuperscript{\dag}}

\newcommand{\fullauthorlist}{%
  \section*{Full Author List}
  \contributorlist{}

  \vspace{0.35em}
  \par\noindent\leadlist{}%
  \thispagestyle{authorfolio}}

\input{sections/00_abstract}

\begin{document}
\raggedbottom
\clubpenalty=10000
\widowpenalty=10000
\makeatletter\@clubpenalty=10000\makeatother
\setlength{\parskip}{0.3\baselineskip}
\setlength{\emergencystretch}{1.5em}
\maketitle
\thispagestyle{titlefolio}
\enlargethispage{0.4cm}

\begingroup
\setlength{\intextsep}{6pt}
\begin{table}[H]
\centering
\small
\setlength{\tabcolsep}{2pt}
\renewcommand{\arraystretch}{1.22}
\caption{\textbf{Accuracy (\%) on the three layers.} Grey tags give each model's reasoning setting; $\pm$ is the standard deviation over three rollouts. Bold marks the best model in a column. A Base item counts as correct only if all its sub-questions are. $^{\dagger}$Excludes items refused by the model's API service.}
\label{tab:rq1-leaderboard}
\begin{tabular*}{\textwidth}{@{\extracolsep{\fill}}l c !{\hspace{0.28em}} cc !{\hspace{0.28em}} ccc@{}}
\toprule
\multirow{2}{*}{\textbf{Model}} & \textbf{Seed} & \multicolumn{2}{c}{\textbf{Base}} & \multicolumn{3}{c}{\textbf{Hard}} \\
\cmidrule(lr){2-2}\cmidrule(lr){3-4}\cmidrule(lr){5-7}
 & Acc. & Item & Sub-Q & Direct & Sandbox & +\,library \\
\midrule
\withlogo{gpt}{GPT-5.6 Sol}\think{max}         & {\acc{\mathbf{89.8}}{3.7}} & {\acc{\mathbf{77.6}}{0.6}} & {\acc{\mathbf{92.6}}{0.2}} & {\acc{13.7}{5.9}} & {\acc{40.3}{4.2}} & {\acc{45.7}{2.1}} \\
\withlogo{claude}{Claude Opus 5}\think{max}       & {\acc{86.5}{0.7}} & {\acc{74.2^{\dagger}}{2.7}} & {\acc{88.9^{\dagger}}{0.7}} & {\acc{\mathbf{15.6}^{\dagger}}{1.8}} & {\acc{\mathbf{55.7}}{3.2}} & {\acc{\mathbf{57.0}}{2.6}} \\
\withlogo{kimi}{Kimi K3}\think{max}             & {\acc{88.6}{0.0}} & {\acc{57.9}{1.8}} & {\acc{81.6}{1.8}} & {\acc{\phantom{0}4.0}{0.0}} & {\acc{41.3}{1.5}} & {\acc{43.0}{8.9}} \\
\withlogo{grok}{Grok 4.6}\think{xhigh}          & {\acc{87.0}{1.8}} & {\acc{52.5}{4.1}} & {\acc{82.9}{0.7}} & {\acc{12.0}{1.0}} & {\acc{39.3}{1.5}} & {\acc{38.3}{2.1}} \\
\withlogo{deepseek}{DeepSeek V4 Pro}\think{max}     & {\acc{83.9}{1.7}} & {\acc{51.8}{1.5}} & {\acc{81.2}{1.2}} & {\acc{\phantom{0}9.3}{1.2}} & {\acc{26.0}{5.6}} & {\acc{20.0}{1.7}} \\
\withlogo{glm}{GLM-5.3}\think{max}             & {\acc{87.3}{0.9}} & {\acc{45.6}{3.2}} & {\acc{67.2}{4.9}} & {\acc{\phantom{0}3.0}{1.0}} & {\acc{20.7}{0.6}} & {\acc{31.3}{5.1}} \\
\withlogo{doubao}{Doubao Seed 2.1 Pro}\think{max} & {\acc{85.8}{0.7}} & {\acc{44.6}{2.2}} & {\acc{73.2}{1.0}} & {\acc{\phantom{0}5.0}{1.0}} & {\acc{26.0}{1.0}} & {\acc{33.7}{0.6}} \\
\bottomrule
\end{tabular*}
\end{table}
\endgroup
\suppressfloats[t]

\begin{figure}[t]
\centering
\includegraphics[width=\textwidth]{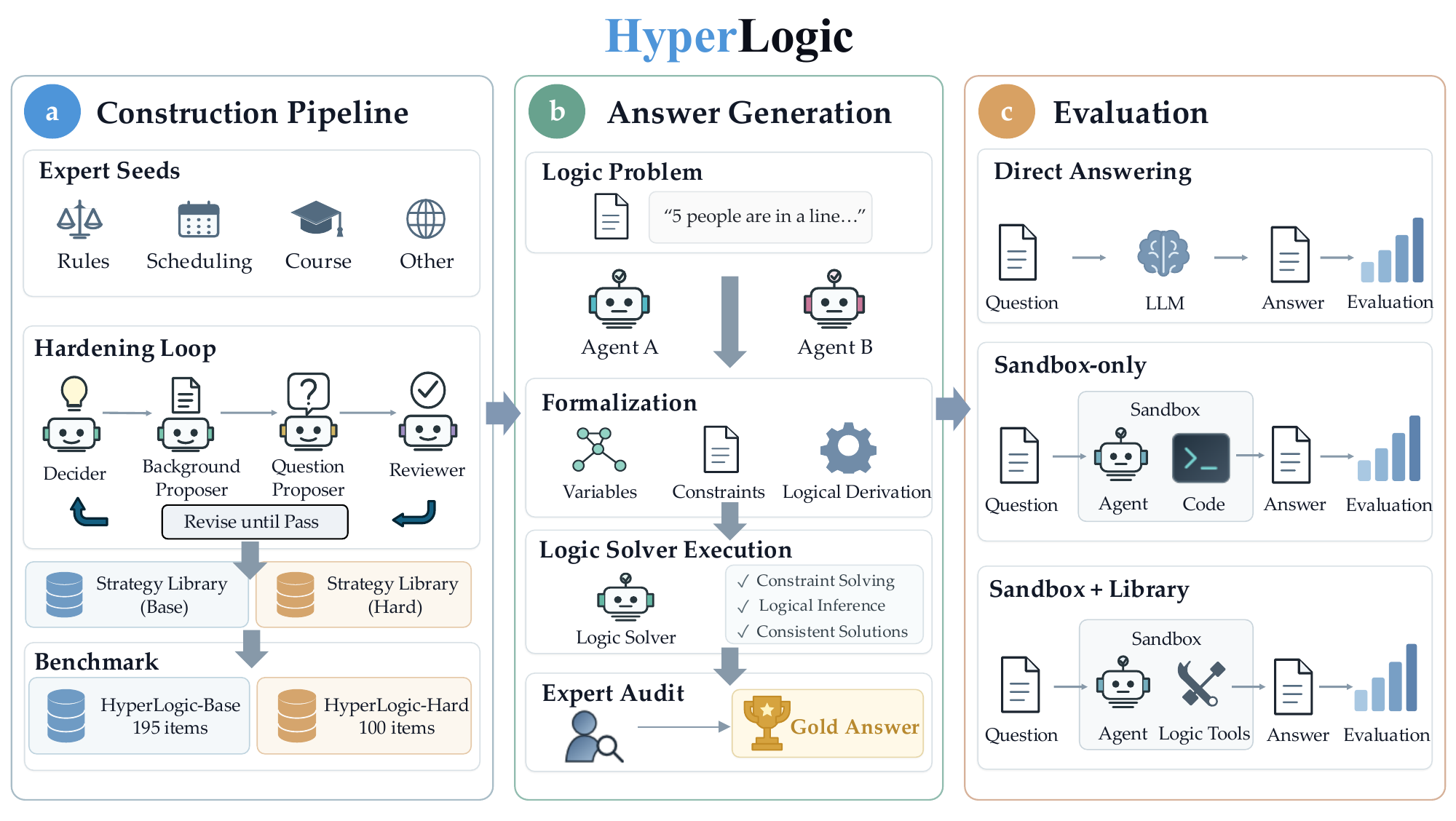}
\caption{\textbf{Overview of HyperLogic.} \textbf{(a)} A four-role agent loop hardens seeds with tier-specific strategy libraries. \textbf{(b)} Two independent formalizations provide solver-derived answers for layered adjudication. \textbf{(c)} Both tiers are evaluated by direct answering; Hard also uses sandbox agents, with or without the logic modeling library.}
\label{fig:overview}
\end{figure}

\input{sections/01_introduction}
\input{sections/02_method}
\input{sections/03_dataset_evaluation}
\input{sections/04_results}
\input{sections/05_related_work}
\input{sections/06_conclusion}
\input{sections/07_ai_use}

\fullauthorlist
\vspace{1.2\baselineskip}

\bibliographystyle{plainnat}
\bibliography{references}

\FloatBarrier
\appendix
\section*{Appendix}
\input{sections/09_appendix_pipeline}
\input{sections/10_appendix_evaluation_prompts}
\input{sections/11_appendix_results}
\clearpage
\input{sections/14_appendix_prompt_figures}
\clearpage
\input{sections/15_appendix_translation_prompt}

\end{document}

%% file: sections/00_abstract.tex
\begin{abstract}
Existing logic benchmarks primarily measure models' ability to answer reasoning questions directly. Scalable benchmarks often generate text from formal structures, which makes answers easy to compute but fixes the formalization before the problem is written. Forward construction preserves the challenge of finding a faithful formalization, yet makes difficulty and answer reliability harder to control. We introduce HyperLogic, a forward-construction pipeline that separates problem authoring from answer generation. A multi-agent workflow hardens undergraduate-authored Chinese seeds without solving them; two agents from different model families independently translate each finished item into executable finite-domain models; their encodings and solver-derived answers undergo layered, agent-assisted adjudication under human-expert oversight. \textbf{HyperLogic-Base} contains $195$ items and $922$ sub-questions and separates seven frontier models by $33.0$ percentage points in strict item accuracy ($44.6$--$77.6\%$). \textbf{HyperLogic-Hard} contains $100$ items with larger, coupled search spaces, on which no model exceeds $16\%$ accuracy in direct answering. We also use Hard to evaluate agents' ability to formalize and solve problems with tools, comparing a code sandbox alone with one that includes our logic modeling library. The sandbox improves every model by $16.7$--$40.1$ points; adding the library helps five models and hurts two. These results highlight the difficulty of faithful formalization even with tool access.
\end{abstract}

%% file: sections/01_introduction.tex
\section{Introduction}
\label{sec:intro}

Logical reasoning over natural language is an important capability of large language models. It is essential in applications such as contract review, regulatory compliance and clinical guideline checking, where a conclusion must follow from the stated premises rather than merely sound plausible~\citep{folio:han2024,logicbench:parmar2024}. Existing benchmarks for this capability are built in two main ways: by generating language from a formal structure, or by writing the problem in natural language first.

Most scalable logic benchmarks use reverse construction: they begin with a formal structure and render it into natural language, including RuleTaker~\citep{ruletaker:clark2020}, ProofWriter~\citep{proofwriter:tafjord2021}, LogicBench~\citep{logicbench:parmar2024}, ZebraLogic~\citep{zebralogic:lin2025}, SATBench~\citep{wei-etal-2025-satbench} and, in Chinese, ChLogic~\citep{chlogic:2026}. This makes both the answer and the difficulty controllable, but it also gives every item a predefined formalization: the structure from which its text was generated. Choosing a faithful model is itself part of the reasoning task, and reverse construction largely removes this step.

We therefore choose forward construction. FOLIO~\citep{folio:han2024} begins with human-written natural-language premises and formalizes them afterwards, so a description may admit several structurally different but answer-equivalent models. This creates two practical problems. First, difficulty is hard to scale: human authors cannot increase logical complexity as cheaply as a symbolic generator, and many existing benchmarks now provide little separation among frontier models~\citep{ouroboros:deveci2025}. Table~\ref{tab:existing-benchmark-saturation} makes the saturation concrete: five of seven benchmarks separate three frontier models by at most three points, and three are already at $97$--$100\%$; HyperLogic-Base opens that spread to $32.0$ points. Second, reliable answers become harder to obtain as the questions become harder. A systematic audit found $42.5\%$ of FOLIO's formalizations incorrect, with corrected labels shifting measured accuracy by $11$ to $23$ points~\citep{fixingfolio:brunello2026}; similar annotation problems have been reported for FrontierMath and Humanity's Last Exam~\citep{frontiermath:glazer2024,hle:phan2025}. More broadly, existing logical-reasoning benchmarks primarily measure direct-answer performance, with relatively little attention to agents' ability to solve logical problems using tools.

\begin{table}[t]
\centering
\small
\setlength{\tabcolsep}{3pt}
\setlength{\databarwidth}{2.2em}
\renewcommand{\arraystretch}{1.18}
\caption{\textbf{Strict accuracy (\%) of three frontier models at maximum reasoning effort, ordered by how far apart they end up.} Existing benchmarks use the same $200$ examples per model; HyperLogic-Base reports item accuracy over its $195$ items, averaged over three rollouts. Missing or unparseable answers count wrong. Spread is the gap between the best and worst model, drawn relative to the largest value in the column.}
\label{tab:existing-benchmark-saturation}
\begin{tabularx}{\textwidth}{@{}Xrrrl@{}}
\toprule
 & \multicolumn{3}{c}{\textbf{Strict accuracy (\%)}} & \\
\cmidrule(lr){2-4}
\textbf{Benchmark} & \modellogo{gpt}\,\textbf{GPT-5.6 Sol} & \modellogo{glm}\,\textbf{GLM-5.3} & \modellogo{deepseek}\,\textbf{DeepSeek V4 Pro} & \textbf{Spread} \\
\midrule
TruthQuest~\citep{truthquest:mondorf2024} & $\mathbf{100.0}$ & $\phantom{0}99.5$ & $\mathbf{100.0}$ & \databar{1.6}{$\phantom{0}0.5$} \\
SATBench~\citep{wei-etal-2025-satbench} & $\phantom{0}97.5$ & $\phantom{0}97.0$ & $\phantom{0}\mathbf{98.0}$ & \databar{3.1}{$\phantom{0}1.0$} \\
FOLIO~\citep{folio:han2024} & $\phantom{0}85.0$ & $\phantom{0}\mathbf{86.0}$ & $\phantom{0}84.5$ & \databar{4.7}{$\phantom{0}1.5$} \\
ZebraLogic~\citep{zebralogic:lin2025} & $\mathbf{100.0}$ & $\phantom{0}99.5$ & $\phantom{0}98.5$ & \databar{4.7}{$\phantom{0}1.5$} \\
DivLogicEval~\citep{chung-etal-2025-divlogiceval} & $\phantom{0}\mathbf{80.5}$ & $\phantom{0}77.5$ & $\phantom{0}79.5$ & \databar{9.4}{$\phantom{0}3.0$} \\
LogicBench~\citep{logicbench:parmar2024} & $\phantom{0}\mathbf{87.0}$ & $\phantom{0}81.0$ & $\phantom{0}77.5$ & \databar{29.7}{$\phantom{0}9.5$} \\
Multi-LogiEval~\citep{multilogieval:patel2024} & $\phantom{0}\mathbf{81.0}$ & $\phantom{0}79.5$ & $\phantom{0}69.0$ & \databar{37.5}{$12.0$} \\
\midrule
\addlinespace[1pt]
\multicolumn{1}{@{}>{\columncolor{casetitlebg}[0pt][\tabcolsep]}l}{\textbf{HyperLogic-Base} (ours)} & \cellcolor{casetitlebg}$\phantom{0}\mathbf{77.6}$ & \cellcolor{casetitlebg}$\phantom{0}45.6$ & \cellcolor{casetitlebg}$\phantom{0}51.8$ & \multicolumn{1}{>{\columncolor{casetitlebg}[\tabcolsep][0pt]}l@{}}{\databar[caseframe]{100}{$\mathbf{32.0}$}} \\
\bottomrule
\end{tabularx}
\end{table}

We address these problems with a single construction pipeline. A hardening agent workflow repeatedly rewrites a seed written by a logic-trained undergraduate to increase its difficulty. Once the item is complete, two agents independently translate it into executable declarative models, and logic solvers derive the answers. The encodings and solver-derived answers undergo layered, agent-assisted adjudication under human-expert oversight; only items whose two corrected encodings yield the same answers are released. This division of labour is what allows the pipeline to scale: the authoring agents write the problem, the formalization agents translate it, and the solvers perform the search. The difficulty of the resulting items can therefore exceed the solving ability of the models used to build them, while their answers remain executable and auditable. Using this pipeline, we construct two datasets. \textbf{HyperLogic-Base} contains problems that frontier models can still answer directly and is designed to separate them. \textbf{HyperLogic-Hard} increases and couples the state space until direct solving is no longer practical, and evaluates whether models can solve such problems by writing code in a sandbox or by using the logic modeling library.

Our contributions are:
\begin{enumerate}[leftmargin=*,itemsep=2pt,topsep=3pt,parsep=0pt]
\item \textbf{A forward construction pipeline that scales difficulty} (Section~\ref{sec:method}). An agent workflow hardens each item; two independent formalizations and logic solvers derive its answer; and layered, agent-assisted adjudication checks fidelity and answer agreement under human-expert oversight before release.
\item \textbf{HyperLogic-Base}, a Chinese logical-reasoning dataset with $195$ items and $922$ sub-questions. It separates seven frontier models by $33.0$ points in item accuracy ($44.6$--$77.6\%$; Section~\ref{sec:rq1}).
\item \textbf{HyperLogic-Hard}, $100$ items with state spaces too large for direct model solving (Section~\ref{sec:rq3}). In direct answering no model exceeds $16\%$; a code sandbox lifts every model by $17$ to $40$ points, while adding the logic modeling library raises five models and lowers two.
\end{enumerate}

%% file: sections/02_method.tex
\section{Forward Construction Pipeline}
\label{sec:method}

HyperLogic is built in four stages (Figure~\ref{fig:overview}). Undergraduates with prior logic coursework write seeds: short, self-contained situations spanning everyday, institutional, and fictional settings (Appendix~\ref{sec:seed-collection}). An agent loop applies Library-Base to seeds to produce Base items, then Library-Hard to selected Base items to produce Hard items, without generating gold answers. Two agents from different model families then independently formalise each finished item, and logic solvers execute their models to derive the answers. Each item then undergoes layered, agent-assisted adjudication centered on the fidelity of its two encodings to the source text and the agreement of their solver-derived answers, under human-expert oversight. With answers derived by solvers instead of the authoring agents, the pipeline can produce items beyond those agents' direct-answering ability.

\subsection{The Hardening Agent Workflow}
\label{sec:method-loop}

A hardening agent workflow rewrites an input item under a selected strategy library. The library specifies the reasoning mechanisms to introduce and the intended difficulty tier; the four agent roles plan, implement, and review those mechanisms. We first apply Library-Base to seeds to produce Base items, then apply Library-Hard to selected Base items to produce Hard items. This authoring loop does not generate gold answers.

The workflow consists of four nodes (Appendix~\ref{sec:hardening-prompts}):
\begin{itemize}[leftmargin=*,itemsep=2pt,topsep=3pt,parsep=0pt]
\item \textbf{Decider.} Reads the input item and selected strategy library, then emits a hardening blueprint: a weakness analysis, a strategy plan for the background, a strategy plan for the questions, and the failure traps the rewritten item is expected to set.
\item \textbf{Background proposer.} Rewrites the situation to that plan and returns, alongside it, a dependency table recording for each new clue why it matters and which sub-questions it feeds.
\item \textbf{Question proposer.} Writes the question set against the locked background, tagging every sub-question with the clues it depends on and the strategies it carries.
\item \textbf{Reviewer.} Sees the candidate \emph{without answers} and judges whether the blueprint was realised. Candidates that fail review are returned to the appropriate node: question-local defects send them back to the question proposer, while defects in the background or the plan send them back to the decider, which turns the findings into concrete fixes. Items that cannot pass within the retry budget are discarded.
\end{itemize}

\subsection{Hardening Strategy Libraries}
\label{sec:method-strategies}

Within the hardening agent workflow, the hardening strategy library is a selectable plug-in: exchanging it changes what kind of item the pipeline emits without touching the workflow itself. A single pipeline can therefore produce different tiers for different purposes, which makes difficulty a configurable property of a construction run rather than a fixed property of the method---a builder who wants a different kind of hard item writes a different strategy library, not a different pipeline.

\paragraph{Library-Base produces items within the models' ceiling.} Its six operators make the reasoning longer: more branches to follow, more distractor premises to rule out, more sub-questions to answer, more aliases to keep track of. The chain grows, but its scale stays inside what a frontier model can carry. The tier it produces is therefore a benchmark of natural-language reasoning within reach: items a frontier model can answer without tools.

\paragraph{Library-Hard produces items past the point of direct solving.} The second library does two things at once: it enlarges the state space past what can be enumerated in context, and it removes the structure that would let that space be cut into independent parts. It retains four Library-Base operators as \emph{carriers} and adds four hardening axes---minimality cores, global coupling constraints, cyclic truth reference, and coupled state dimensions---each with four implementations; Appendices~\ref{sec:strategy-libraries} and~\ref{sec:strategy-library-hard} list both libraries in full. The tier this produces is a benchmark of reasoning beyond direct solving, which is why Hard is evaluated under tool conditions (Section~\ref{sec:rq3}) rather than by direct answering alone.

\subsection{Answer Generation}
\label{sec:method-decouple}

To obtain auditable gold answers without requiring the authoring agents to solve the items, we translate each item into a finite-domain theory and evaluate its queries with logic solvers.

To make these translations uniform and auditable, we give every formalization agent the same logic modeling library. Each agent uses it to encode the background as a finite-domain theory $T=(V,\Delta,\Gamma)$: free variables $V$ with closed finite domains; derived terms $\Delta$, whose values are determined by $V$ and introduce no independent choices; and constraints $\Gamma$, each citing the background claims that license it. A world assigns each variable in $V$ a value from its domain; $\mathrm{Mod}(T)$ contains the worlds in which every constraint is defined and true using the definitions in $\Delta$.

Each sub-question is expressed as a query over the admissible worlds in a specified scope $s$. A scope may add assumptions or replace or retract named constraints, yielding a theory $T_s$ and model set $M_s=\mathrm{Mod}(T_s)$; baseline questions use $T_s=T$. The library supports six query modalities---possibility ($\exists w\in M_s\,\varphi(w)$), necessity ($\forall w\in M_s\,\varphi(w)$), uniqueness, enumeration, exact counting, and incompatibility. The modal formulas assume nonempty scopes and well-defined propositions. Each operation is dispatched by capability to SAT (CaDiCaL~\citep{cadical:biere2024}), SMT (Z3~\citep{demoura2008z3}), BDDs~\citep{bdd:bryant1986}, or model counting (Ganak~\citep{ganak:soos2025}). Answers are computed under the scopes specified by the question. Appendix~\ref{sec:formalisation-interface} gives the primitives, query semantics, boundary cases, and solver dispatch.

Two frontier agents---GPT-5.6 Sol~\citep{gpt56:openai2026} under Codex~\citep{codex:openai2026} and Claude Opus 5~\citep{claudeopus5:anthropic2026} under Claude Code~\citep{claudecode:anthropic2026}---each receive the same item, the same skill and the same logic modeling library, and independently produce a translation together with its solver-derived answers. The encodings, the answers, and the agents' authoring traces are the inputs to expert audit (Section~\ref{sec:method-audit}).

\subsection{Expert Audit}
\label{sec:method-audit}

Before accepting solver-derived answers as gold, we audit the encodings against the original background and questions. Two independently produced encodings can share a misreading even when their answers agree. Automated checks identify potential defects, and a human logic expert reviews the evidence and resolves discrepancies with agent assistance.

\paragraph{Coverage.} Each substantive statement in the background is linked to the variables, derived terms, or constraints that represent it. Statements that impose no logical condition receive an explicit explanation. Automated checks flag missing coverage and invalid references to source statements.

\paragraph{Consistency with the source text.} We check for both missing rules and unsupported restrictions. To detect missing rules, the checker examines concrete worlds admitted by an encoding and identifies any that violate the background. To detect unsupported restrictions, individual constraints are temporarily removed, and newly admitted worlds are checked against the source text. A newly admitted world that the background permits indicates that the removed constraint excludes a valid situation. These checks provide concrete cases and source statements for expert review.

\paragraph{Expert adjudication.} The expert examines both encodings, their answers, the translation traces, and the available check reports. An assisting agent helps locate discrepancies and prepare repairs. The review also examines agreeing answers for shared misinterpretations. Faulty encodings are corrected and rerun; ambiguous items are clarified or excluded. Release requires matching substantive answers from the two reviewed encodings. Appendix~\ref{sec:audit-records} provides the detailed audit procedures and completed adjudication records.

%% file: sections/03_dataset_evaluation.tex
\section{Dataset and Evaluation}
\label{sec:datasets}

\subsection{Dataset}

Using the forward construction pipeline described in Section~\ref{sec:method}, we construct HyperLogic from $228$ single-question seeds contributed by undergraduates with prior coursework in logic. We first harden these seeds with Library-Base, obtaining $195$ Base items containing $922$ sub-questions. Each Base item pairs a shared background with multiple sub-questions, averaging $4.7$ sub-questions per item and approximately $3{,}300$ Chinese characters per background. We then apply Library-Hard to further harden Base items, producing Hard: $100$ single-question items with backgrounds averaging approximately $8{,}700$ Chinese characters.

\subsection{Evaluation}

We evaluate seven models: GPT-5.6 Sol~\citep{gpt56:openai2026}, Claude Opus 5~\citep{claudeopus5:anthropic2026}, Kimi K3~\citep{kimik3:2026}, Grok 4.6~\citep{grok46:xai2026}, DeepSeek V4 Pro~\citep{deepseekv4:2026,deepseekv4pro:2026}, GLM-5.3~\citep{glm53:zai2026}, and Doubao Seed 2.1 Pro~\citep{seed21:bytedance2026}. Each model runs at the highest reasoning setting supported by its evaluation route.

We consider three evaluation conditions. \textbf{Direct answering} uses a single model call without tools. The \textbf{sandbox-only agent} operates in a networked sandbox with shell access, where it can write code and import solvers. The \textbf{library-available agent} additionally receives the logic modeling library used for answer generation in Section~\ref{sec:method-decouple} and its manual. Use of the supplied library is optional. Agent evaluations use Codex for GPT-5.6 Sol and Claude Code for the other models, with two CPU cores and $10$\,GB of memory per sandbox. Each attempt has a three-hour wall-clock budget across all models and conditions. Seed and Base are evaluated under direct answering, while Hard is evaluated under all three conditions.

\begin{table}[!t]
\centering
\small
\setlength{\tabcolsep}{3pt}
\setlength{\databarwidth}{2.7em}
\renewcommand{\arraystretch}{1.15}
\newcommand{\tokc}[1]{\textcolor{ratiolow}{\textbf{#1}}}
\newcommand{\tokw}[1]{\textcolor{ratiohigh}{\textbf{#1}}}
\caption{\textbf{Median reasoning tokens on Base over three rollouts.} The three difficulty groups contain 65 items each and are shared across models. Within each group, the larger of the correct and incorrect medians is marked green (correct) or red (incorrect).}
\label{tab:tokens}
\begin{tabular*}{\textwidth}{@{\extracolsep{\fill}}lrlrrrrrr@{}}
\toprule
\multirow{2}{*}{\textbf{Model}} & \multirow{2}{*}{\textbf{Base acc.}} & \multirow{2}{*}{\textbf{All items}} & \multicolumn{2}{c}{\textbf{Easiest third}} & \multicolumn{2}{c}{\textbf{Middle third}} & \multicolumn{2}{c}{\textbf{Hardest third}} \\
\cmidrule(lr){4-5}\cmidrule(lr){6-7}\cmidrule(l){8-9}
 & & & Correct & Wrong & Correct & Wrong & Correct & Wrong \\
\midrule
\withlogo{gpt}{GPT-5.6 Sol} & $77.6$ & \databar{42.3}{22.3k} & 11.4k & \tokw{12.9k} & \tokc{26.9k} & 13.9k & \tokc{31.9k} & 18.6k \\
\withlogo{claude}{Claude Opus 5} & $74.2$ & \databar{30.5}{16.1k} & \tokc{7.4k} & 3.5k & \tokc{63.1k} & 6.5k & \tokc{73.4k} & 9.5k \\
\withlogo{kimi}{Kimi K3} & $57.9$ & \databar{46.2}{24.3k} & 13.5k & \tokw{22.9k} & \tokc{45.8k} & 28.0k & \tokc{43.0k} & 30.6k \\
\withlogo{grok}{Grok 4.6} & $52.5$ & \databar{38.0}{20.0k} & 11.3k & \tokw{15.5k} & \tokc{35.7k} & 28.0k & \tokc{33.2k} & 25.0k \\
\withlogo{deepseek}{DeepSeek V4 Pro} & $51.8$ & \databar{75.2}{39.6k} & 28.3k & \tokw{34.3k} & \tokc{57.3k} & 44.9k & \tokc{61.6k} & 41.4k \\
\withlogo{glm}{GLM-5.3} & $45.6$ & \databar{100}{52.7k} & 32.3k & \tokw{47.4k} & 52.8k & \tokw{61.8k} & 63.7k & \tokw{67.8k} \\
\withlogo{doubao}{Doubao Seed 2.1 Pro} & $44.6$ & \databar{28.8}{15.2k} & 6.5k & \tokw{20.4k} & 14.5k & \tokw{31.2k} & 11.9k & \tokw{26.4k} \\
\bottomrule
\end{tabular*}
\end{table}

DeepSeek V4 Pro judges each answer against its reference for semantic equivalence, allowing harmless wording and ordering differences. Appendix~\ref{sec:evaluation-prompts} provides the agent evaluation and judging prompts. For Base, we report both sub-question accuracy and item accuracy; an item is correct only when all its sub-questions are correct. Unanswered items count as incorrect. We run each model three times under each applicable condition and report the mean accuracy and standard deviation across rollouts in Table~\ref{tab:rq1-leaderboard}.

%% file: sections/04_results.tex
\section{Results and Findings}
\label{sec:results}
\FloatBarrier

\subsection{Base: Separation and Failure Modes}
\label{sec:rq1}
\label{sec:findings-tokens}

\finding{1}{Base is a hard and discriminative benchmark.} As shown in Table~\ref{tab:existing-benchmark-saturation}, existing natural-language reasoning benchmarks barely separate frontier models: on five of the seven, three frontier models differ by at most three points. Our Seed layer behaves the same way, with all seven models within $5.9$ points of each other ($83.9$--$89.8\%$). On HyperLogic-Base, by contrast, every model loses at least $12$ points from Seed and the seven models spread over $33.0$ points ($44.6$--$77.6\%$), making Base both hard and discriminative even for state-of-the-art models. Among the seven models we test, GPT-5.6 Sol and Claude Opus 5 rank well above the rest (Table~\ref{tab:rq1-leaderboard}): both reach about $75\%$ ($77.6\%$ and $74.2\%$), at least $16.3$ points ahead of the third model, Kimi K3. When they do fail, the failure is usually narrow: $70\%$ of GPT-5.6 Sol's failed items and $56\%$ of Claude Opus 5's get only one sub-question wrong.

\finding{2}{Longer reasoning does not consistently correspond to higher accuracy.} To compare reasoning length across difficulty levels, we rank Base items by the seven models' mean error rate over three rollouts and divide them into three shared groups of $65$ (Table~\ref{tab:tokens}). Median reasoning length varies substantially across models, from $15.2$k tokens per item for Doubao Seed 2.1 Pro to $52.7$k for GLM-5.3. These differences do not align with accuracy: the two most accurate models, GPT-5.6 Sol and Claude Opus 5, have medians of $22.3$k and $16.1$k tokens, respectively, both less than half that of GLM-5.3.

DeepSeek V4 Pro and GLM-5.3 stand out for their particularly long reasoning traces, even on easy items. On items with a mean error rate of at most $20\%$, their medians are $27.3$k and $27.9$k tokens, compared with $5.6$k--$12.5$k for the other models. These models tend to repeatedly check and verify their reasoning, revisiting their calculations and regenerating their answers. In completed runs, they generate an estimated $6.4$k and $6.8$k tokens on average after the first complete answer draft, about nine times as much as Kimi K3. Figures~\ref{legacy:fig:trace-ds} and~\ref{legacy:fig:trace-glm} show examples of reasoning traces in which complete answers are generated multiple times. Moreover, GLM-5.3 reaches its $131$k-token output limit without a parseable answer in $99$ of its $585$ runs.

Within the shared difficulty groups, the relationship between reasoning length and correctness varies by model. Claude Opus 5 has longer median traces for correct answers in all three groups, with the largest gap in the hardest group: $73.4$k tokens for correct answers versus $9.5$k for incorrect ones. GLM-5.3 and Doubao Seed 2.1 Pro show the reverse in every group. For GPT-5.6 Sol, Kimi K3, Grok 4.6, and DeepSeek V4 Pro, the direction changes with difficulty: incorrect answers have longer median traces in the easiest group, while correct answers have longer median traces in the middle and hardest groups.

\begin{figure}[t]
\centering
\includegraphics[width=\textwidth]{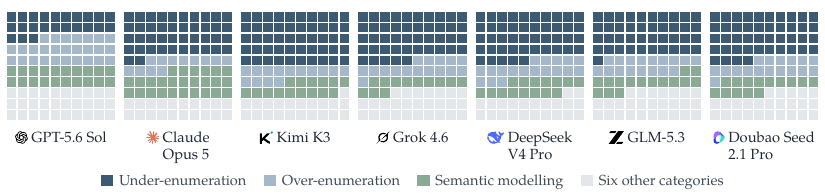}
\caption{\textbf{Base error distributions by model over three rollouts, classified by DeepSeek V4.1 Flash.} Each grid shows one model's classified errors, one square per percent, grouped into under-enumeration, over-enumeration, semantic modelling and the six other categories.}
\label{fig:base-errors}
\end{figure}

\finding{3}{Error sources are shared across models.} To analyze model failures, we use DeepSeek V4.1 Flash~\citep{deepseekv41flash:2026} to classify each incorrect answer into one of nine categories, including under-enumeration, over-enumeration, and semantic modelling (Figure~\ref{fig:base-errors}). Appendix~\ref{legacy:sec:base-error-taxonomy} defines all nine categories and reports their per-model distributions.

Among the nine categories, enumeration errors dominate. For every model, under- and over-enumeration together account for most of its errors, between $50\%$ and $64\%$. The two are not balanced: except for GPT-5.6 Sol, whose errors split almost evenly between them ($23\%$ and $27\%$), every model misses valid cases $2.4$ to $3.6$ times as often as it adds invalid ones, which suggests a tendency to rule out possibilities too early. Semantic modelling, where the model misreads a rule or condition, is the next largest category, at $12\%$ to $26\%$ of each model's errors. The remaining six categories, including confusions between modal and quantifier terms, are minor: none exceeds $10\%$ for any model. Taken together, these results show that current models are far weaker at tracking every possibility the rules allow than at reading the rules correctly.

The size of the possibility space also separates the models. We split enumeration and counting sub-questions at the median size of their gold answer: more than $6$ listed cases, or a count above $12$, counts as large (Figure~\ref{fig:answer-size}). Enumeration size counts listed cases, whether tuples, sets, or members; counting size is the reported integer. For most models, errors rise with size. GLM-5.3's error rate on the larger half is $2.3$ times its rate on the smaller half, and Doubao Seed 2.1 Pro's is $1.9$ times; Kimi K3 ($1.5$) and Grok 4.6 ($1.4$) show a smaller rise, and DeepSeek V4 Pro almost none. The two leaders move the other way: Claude Opus 5 and GPT-5.6 Sol err less on the larger half, at $0.7$ and $0.5$ times their rate on the smaller one. They hold up as the space grows, whereas the other models struggle once the possibility space becomes large.

\subsection{Hard: Direct Answering and Agent Sandboxes}
\label{sec:rq3}

\finding{4}{Hard is largely beyond direct answering; an agent sandbox helps but does not solve it.} No model exceeds $16\%$ accuracy when answering Hard directly. Giving models a code sandbox improves every model, both without and with the logic modeling library: gains over direct answering range from $16.7$ to $40.1$ percentage points in the sandbox-only condition and from $10.7$ to $41.4$ points in the library-available condition. The highest accuracies reach only $55.7\%$ and $57.0\%$, respectively (Table~\ref{tab:rq1-leaderboard}). Current models remain far from reliable on Hard's large, multidimensional, and tightly coupled search spaces despite access to solvers. Among the hardening strategies used in Hard, minimal explanation cores are associated with the lowest sandbox accuracy, only $15.3\%$. This suggests that models struggle to trace how constraints jointly cause a contradiction and verify that every constraint in the selected core is necessary. Appendix~\ref{sec:strategy-accuracy} reports accuracy by Hard strategy under both agent conditions.

\finding{5}{The logic modeling library does not consistently improve accuracy.} Adding the library to the agent sandbox changes Hard accuracy by $-6.0$ to $+10.6$ percentage points across seven models: five improve and two decline (Table~\ref{tab:rq1-leaderboard}). Making the library available therefore does not guarantee a better answer. Using it effectively also requires the model to understand the library's modeling language and express its reasoning through that interface.

\begin{figure}[t]
\begin{minipage}[t]{0.48\textwidth}
\centering
\includegraphics[width=\linewidth]{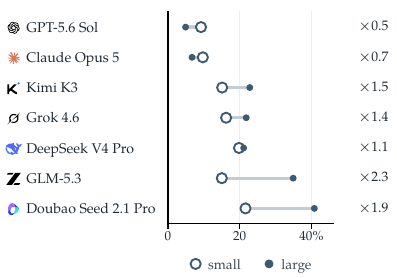}
\caption{\textbf{Error rates (\%) on enumeration and counting sub-questions with small (at or below the median) and large (above the median) gold answers.} Numbers on the right give the ratio of error rates for large versus small answers.}
\label{fig:answer-size}
\end{minipage}\hfill
\begin{minipage}[t]{0.48\textwidth}
\centering
\includegraphics[width=\linewidth]{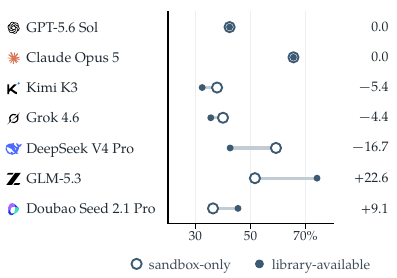}
\caption{\textbf{Accuracy (\%) on matched item--rollout pairs where both conditions return a scored answer and the library-available workspace contains logic modeling library code.} Numbers on the right show the change from sandbox-only to library-available in percentage points.}
\label{fig:tool-paired}
\end{minipage}
\end{figure}

To examine this variation, Figure~\ref{fig:tool-paired} matches the same item and rollout across the two agent conditions, retaining pairs in which both conditions produce a scored answer and the library-available workspace contains library code. On these selected pairs, GLM-5.3 and Doubao improve; GPT-5.6 Sol and Claude Opus 5 remain unchanged; and Kimi K3, Grok 4.6, and DeepSeek V4 Pro decline. The largest opposing changes are GLM-5.3's increase from $51.6\%$ to $74.2\%$ and DeepSeek V4 Pro's decrease from $59.3\%$ to $42.6\%$. These opposing changes suggest differences in how models handle the library's interface: they must choose suitable modeling primitives and operations, encode the source rules in the required syntax, and formulate the query it will solve. If an agent chooses an unsuitable operation or mistranslates a rule, the solver can return a precise but incorrect answer to the original question; the DeepSeek V4 Pro example in Figure~\ref{fig:encoding-case} illustrates this outcome.

\begin{figure}[!t]
\centering
\includegraphics[width=\textwidth]{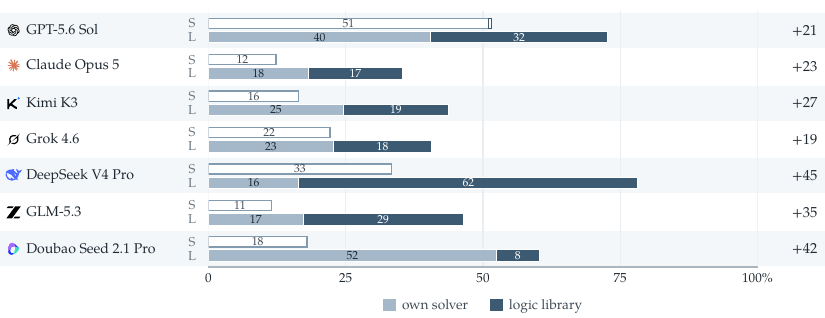}
\caption{\textbf{Computational routes among agent attempts with saved code.} S denotes sandbox-only (outlined bars), and L denotes library-available (filled bars). Segments show independently imported solver code (left) and logic modeling library code (right); the remaining share is plain Python. The right column reports the increase in the combined solver share from S to L, in percentage points. Attempts without saved code are excluded.}
\label{fig:tool-mix}
\end{figure}

\finding{6}{Models use the sandbox through different computational routes.} Figure~\ref{fig:tool-mix} groups attempts with saved code into three routes: code referencing the logic modeling library, code importing an independent symbolic solver, and plain Python code with neither detected. In the sandbox-only condition, six of seven models predominantly write plain Python. For five models, it accounts for $78\%$--$89\%$ of coded attempts; DeepSeek V4 Pro uses it in $67\%$, while GPT-5.6 Sol splits almost evenly between plain Python and an imported solver. These Python programs commonly enumerate candidate assignments and filter them with handwritten rules. Access to a sandbox alone therefore seldom leads models to choose a symbolic solver.

When the logic modeling library and its manual are available, the share of coded attempts containing either library code or an independently imported solver rises by $19$--$45$ percentage points for every model. Their choices still differ: DeepSeek V4 Pro has library code in $62\%$ of coded attempts, whereas Doubao Seed 2.1 Pro has it in only $8\%$ and imports its own solver in $52\%$. Claude Opus 5, Kimi K3, Grok 4.6, and GLM-5.3 remain predominantly with plain Python, while GPT-5.6 Sol uses all three routes. Doubao's independently imported solver share rises from $18\%$ to $52\%$ despite its limited use of the supplied interface. This pattern suggests that the library and manual can serve as a modeling reference, encouraging solver-backed approaches even when a model chooses another interface.

%% file: sections/05_related_work.tex
\section{Related Work}
\label{sec:related}

Natural-language logic benchmarks can use reverse construction, generating text from predefined rules or constraints. LogicBench builds scenarios around inference rules~\citep{logicbench:parmar2024}, and SATBench translates SAT formulas into puzzles~\citep{wei-etal-2025-satbench}. This makes answers and difficulty controllable, while fixing the intended formalization before the text is written. Forward construction starts from natural-language problems: ReClor and LogiQA collect human-authored examination questions~\citep{reclor:yu2020,logiqa:liu2020}, while FOLIO's WikiLogic subset pairs human-written scenarios with logical annotations~\citep{folio:han2024}. This preserves the challenge of finding a faithful formalization, but scaling requires further authoring and answer validation.

To increase the scale and difficulty of such problems, recent work automatically transforms existing items. LogiHard combines answer options using logical operators and derives new labels from the original answers~\citep{logihard:liu2026}; AutoLogi converts questions into open-ended puzzles, varies constraints, and verifies solutions programmatically~\citep{autologi:zhu2025}. However, these transformations operate within the original option set or variable domains, limiting their scope for increasing the underlying state space by orders of magnitude. HyperLogic introduces a forward construction pipeline that starts from human-authored natural-language seeds and rewrites their backgrounds and questions using different strategy libraries for Base and Hard. Authoring is separated from answer generation: independent formalizations, solver execution, and expert review establish answers after each item is finalized, allowing difficulty to exceed the authoring models' ability to answer directly.

Existing logical reasoning evaluations primarily focus on models generating answers directly, as in ZebraLogic and AutoLogi~\citep{zebralogic:lin2025,autologi:zhu2025}. Logic-LM studies formalization and symbolic solving within a prescribed pipeline~\citep{logiclm:pan2023}, while autonomous solving in general-purpose sandboxes remains less explored. As difficulty increases, state spaces and computational demands can exceed models' capacity for direct reasoning, making the ability to solve problems with tools an important dimension of evaluation. HyperLogic-Hard compares direct answering, a code sandbox, and a sandbox with an optional logic modeling library on the same items, testing whether tool access translates into effective modeling and solving.

%% file: sections/06_conclusion.tex
\section{Conclusion}
\label{sec:conclusion}

We introduced HyperLogic, a logical-reasoning benchmark built through a forward construction pipeline that separates problem authoring from answer generation. Independent formalizations, logic solvers, and adjudication under human-expert oversight provide auditable answers to items that can exceed the authoring models' direct-solving ability. HyperLogic-Base separates seven frontier models under direct answering, while HyperLogic-Hard evaluates their ability to solve more demanding problems both directly and with tools. On Hard, no model exceeds $16\%$ accuracy in direct answering; code sandboxes improve every model, but the best agent accuracy remains $57.0\%$, and adding a logic modeling library does not consistently help. These results reveal persistent limitations in models' ability to reason logically over natural language and to solve highly challenging logical problems using tools.

%% file: sections/07_ai_use.tex
\section*{AI Use Statement}
\label{sec:ai-use}

We used generative AI tools for benchmark construction and evaluation, including hardening human-authored seeds, translating items into formal models, assisting with auditing and repairs, judging model responses, and categorizing errors. AI tools also assisted with methodological feedback, checking formal definitions, implementing and debugging code, analyzing and interpreting results, preparing figures, and drafting and revising the manuscript. Reference answers are computed by logic solvers and checked through the audit procedure under human-expert oversight described in Section~\ref{sec:method-audit}. The authors retain responsibility for the benchmark, the reported results and claims, and the final manuscript.

%% file: sections/09_appendix_pipeline.tex
\section{Forward Construction Pipeline}
\label{sec:forward-construction-appendix}

\subsection{Seed collection}
\label{sec:seed-collection}

We collected $228$ original, single-question Chinese logic problems from undergraduate contributors with prior coursework in logic. Each seed describes a short, self-contained situation in natural language. The settings span everyday life, institutional rules, technical systems, and fictional puzzles, including travel planning, course selection, recruitment, security alerts, and knights and knaves.

\subsection{Hardening prompt templates}
\label{sec:hardening-prompts}

Figures~\ref{fig:prompt-decider}--\ref{fig:prompt-reviewer} present the hardening prompt templates for the four roles in the authoring workflow: decider, background proposer, question proposer, and reviewer. The strategy-library placeholders in these templates can be filled with the content of the selected library.

\subsection{Library-Base: six authoring operators}
\label{sec:strategy-libraries}

Library-Base uses six operators to increase the number of live cases and the amount of bookkeeping in a natural-language item.

\paragraph{Branching (\texttt{add\_branching}).} Replace a single chain of inference with multiple branches, intersecting conditions, or alternatives that require backtracking and comparison. For example, alternative eligibility routes may depend on overlapping time windows and evidence conditions.

\paragraph{Distractor premise (\texttt{add\_distractor\_premise}).} Add information that closely resembles decisive evidence in its wording, entities, or context but does not itself determine the answer. Resolving the problem then involves distinguishing relevant evidence from plausible distractions.

\paragraph{Richer answer shape (\texttt{change\_question\_to\_set\_output}).} Expand a simple question into an array of independent subquestions with richer outputs suited to the problem structure. These can ask for feasible cases, projected sets, exact counts, uniqueness, or the existence of alternatives.

\paragraph{Genuine uncertainty (\texttt{add\_uncertainty\_or\_multi\_answer}).} Introduce several compatible outcomes, a situation whose result is not uniquely determined, or a question asking for multiple feasible answers. The solver must retain the possibilities allowed by the evidence instead of prematurely selecting one.

\paragraph{Counterfactual variant (\texttt{add\_counterfactual\_variant}).} Keep the shared baseline background fixed and introduce separate subquestions that flip different key facts. Comparing the baseline with these hypothetical situations requires recomputing their consequences.

\paragraph{Alias and coreference variation (\texttt{alias\_and\_coreference\_variation}).} Refer to the same entities through alternative names, abbreviations, descriptive phrases, or coreferential expressions. The solver must track these identities across the background and questions.

\subsection{Library-Hard: four depth axes}
\label{sec:strategy-library-hard}

{\sloppy Hard retains \texttt{add\_\allowbreak{}branching}, \texttt{add\_\allowbreak{}distractor\_\allowbreak{}premise}, \texttt{change\_\allowbreak{}question\_\allowbreak{}to\_\allowbreak{}set\_\allowbreak{}output}, and \texttt{add\_\allowbreak{}uncertainty\_\allowbreak{}or\_\allowbreak{}multi\_\allowbreak{}answer} as carriers, with the meanings given in Appendix~\ref{sec:strategy-libraries}. It adds the sixteen implementations below, grouped into four depth axes, for twenty entries in total. Counterfactual variants and alias variation are absent from the Hard catalogue.\par}

\paragraph{Minimality core (\texttt{minimality\_core}).} Search for a minimal set of evidence, conditions, or changes that achieves a specified property. Inclusion-minimality means that no proper subset achieves that property; minimum cardinality means having the fewest members among all qualifying sets.
\begin{itemize}[leftmargin=*,itemsep=2pt,topsep=2pt]
\item \emph{Minimal sufficient set}: find an inclusion-minimal subset of named evidence that entails a target conclusion; removing any selected member makes the remaining evidence insufficient.
\item \emph{Minimal hitting set}: find an inclusion-minimal selection intersecting every named conflict or requirement set.
\item \emph{Minimal explanation core}: find a minimally unsatisfiable subset of conflicting rules or records. Its premises are jointly inconsistent, but removing any one restores satisfiability.
\item \emph{Minimal repair set}: identify the fewest permitted changes that make a currently false target hold.
\end{itemize}

\paragraph{Global coupling constraint (\texttt{global\_coupling\_constraint}).} Link multiple state axes through a shared constraint, so their feasible assignments depend on one another.
\begin{itemize}[leftmargin=*,itemsep=2pt,topsep=2pt]
\item \emph{Cardinality quota}: impose a quota such as exactly $k$ eligible people or satisfied conditions, where the counted predicate depends on multiple state axes.
\item \emph{Statement-truth quota}: constrain the number of complete statements that are true. Each statement combines multiple assertions, whose joint evaluation determines the truth of the whole statement.
\item \emph{Conservation balance}: impose conservation or flow balance, such as equal inflow and outflow or allocations summing to a fixed total, so a change propagates across entities and state axes.
\item \emph{Global unsatisfiable core}: make the full rule set jointly unsatisfiable through the interaction of multiple rules, yielding zero configurations that satisfy all rules.
\end{itemize}

\paragraph{Cyclic truth reference (\texttt{cyclic\_truth\_reference}).} Couple truth values with the attributes or other truth values on which they depend, forming a cycle that must be resolved as a whole.
\begin{itemize}[leftmargin=*,itemsep=2pt,topsep=2pt]
\item \emph{Identity-type cycle}: characters have unknown truth-teller or liar types, and statements refer cyclically to other characters' types or the truth of their statements.
\item \emph{Authority-truth cycle}: a statement's truth depends on whether its speaker has the relevant qualification or permission at that time, while those qualifications are themselves constrained by the statements' contents.
\item \emph{Negation cycle}: rules of the form ``this holds unless another rule holds'' create cyclic dependencies through negation, with admissible configurations defined by stable-model semantics.
\item \emph{Recursive fixed-point cascade}: define validity recursively through references from valid entries, so the valid set is determined by a closed fixed point of the recursive definition.
\end{itemize}

\paragraph{Coupled state dimensions (\texttt{coupled\_state\_dimensions}).} Give each entity multiple state fields and couple them across fields and entities, requiring reasoning over their joint assignments.
\begin{itemize}[leftmargin=*,itemsep=2pt,topsep=2pt]
\item \emph{Multi-attribute lock}: assign each entity several attributes and impose constraints both across attributes and across entities.
\item \emph{Time-state transition}: use state-transition rules and time windows to couple an entity's states across time, allowing constraints on one time point to propagate forwards and backwards.
\item \emph{Role-permission validity}: couple roles, permissions, and validity periods. Permission depends jointly on the role and whether the relevant validity period includes the time in question.
\item \emph{Serial dependency chain}: make each decision depend on an earlier one, imposing an order of reasoning in which intermediate choices constrain later conclusions.
\end{itemize}

\section{Answer Generation}
\label{sec:formalisation-interface}

\subsection{Finite-domain theory and modelling interface}

For a finished item, an independently working translator writes a Python \texttt{model.py} whose \texttt{build()} constructs $T=(V,\Delta,\Gamma)$. The method \texttt{var(name, domain, alias=...)} declares a variable with a closed finite domain and a readable alias in the background's vocabulary. The method \texttt{derive(name, expr, alias=..., cites=...)} records an expression whose value, when defined, is fixed by the free variables; it introduces no independent choice. The method \texttt{constrain(expr, id=..., cites=...)} asserts a named rule. Its citations identify the background claims that justify it in a sentence-level census. One independently meaningful assertion should be one named constraint, so a counterfactual can replace it and an audit can remove it without deleting unrelated rules.

Under-evidenced Boolean facts remain two-valued: uncertainty is represented by multiple admissible worlds, rather than an additional ``unknown'' value. A genuinely inapplicable event may instead have a declared not-applicable domain value. Derived terms require explicit finite output domains when the library cannot infer them. Partial operations remain partial: \texttt{select\_unique} returns its declared default when no candidate holds, a label when exactly one holds, and is undefined when several hold. These distinctions matter to uniqueness and counting questions.

The expression library contains Boolean connectives (including implication, biconditional, and exclusive disjunction), equality and finite membership, comparisons and order ranks, all-different and compatibility tables, conditional terms, finite sums, and cardinality operators \texttt{exactly\_k}, \texttt{at\_least\_k}, and \texttt{at\_most\_k}. A complete world assigns every variable in $V$ a value from its declared domain; it belongs to $\mathrm{Mod}(T)$ when all constraints in $\Gamma$ are defined and true using the definitions in $\Delta$. Derived terms used only in queries do not impose additional background constraints; structural checks flag undefined derived values for repair. Different internal encodings must preserve the background's definition of a complete world for world counts and the requested output representation for answer comparison.

\subsection{Questions as queries over worlds}

The companion \texttt{asks.py} translates each verbatim question into a query with one recorded modality and return shape. In a scope $s$, queries operate on $M_s=\mathrm{Mod}(T_s)$, where $T_s$ is the theory after applying that scope. Enumeration returns rows or a set of explicitly named members; \texttt{possible} asks whether a witness world exists, \texttt{necessary} whether a nonempty set of worlds all satisfy a proposition, \texttt{is\_unique} whether the requested value takes exactly one value across those worlds, \texttt{count} returns an exact integer, and \texttt{incompatible\_cases} identifies named partial assignments with no completion. Modal questions over a family use \texttt{possible\_set} or \texttt{necessary\_set} to test each member's predicate. By default, \texttt{rows} preserves one projected row per world (\texttt{dedup=False}), whereas \texttt{count} counts distinct projected rows (\texttt{dedup=True}); setting \texttt{dedup=False} for \texttt{count} counts complete worlds.

For a proposition $\varphi$ defined throughout $M_s$, possibility means $\exists w\in M_s\,\varphi(w)$, while necessity means $M_s\ne\varnothing\land\forall w\in M_s\,\varphi(w)$. The library uses a non-vacuous convention: an empty scope returns false for both modal queries. For a value expression $f$ defined throughout the scope, uniqueness means $|\{f(w):w\in M_s\}|=1$. These semantics require the relevant expressions to be defined on every admissible world; an undefined derived value is a defect to repair, not grounds for silently excluding that world.

Counterfactuals and evidence-restricted questions operate on a \emph{scope}. A scope can assume a hypothesis, replace or retract a named constraint, disable a named rule used as a derivation route, or restrict available evidence by tag or identifier. Tag filtering retains untagged constraints, whereas identifier filtering selects only the named constraints; scope assumptions remain active in both cases. Disabling a rule does not itself assert or retract its premise; any premise stated independently remains governed by the other constraints. For a counterfactual that explicitly changes a baseline fact or rule, the encoding replaces or retracts the original before imposing the new condition; conjoining contradictory conditions would instead make the scope inconsistent. Per-case results remain labelled. Row differences and intersections require the same ordered fields on both sides; row differences preserve duplicates unless explicitly deduplicated.

Under default routing, small declared assignment spaces use exhaustive search, also the semantic oracle for cross-checks. For larger spaces, decision queries and distinct-projection enumeration use SAT (CaDiCaL~\citep{cadical:biere2024} through PySAT~\citep{pysat:ignatiev2018}), witness sampling and enumeration without deduplication use Z3, and complete-world counts try Ganak followed by a BDD backend built on the \texttt{dd} package~\citep{dd:tulip2023}, with fallbacks when an engine is unavailable or cannot serve the operation. Ganak uses probabilistic component caching, so its exact-count guarantee is probabilistic~\citep{ganak:soos2025}. Count results record exactness. The integer-returning \texttt{count} interface rejects inexact results, and \texttt{rows} rejects truncated enumeration. Incomplete results must not be used as gold answers.

\subsection{Translation Skill}

Both formalization agents follow the same translation skill. They record source claims and modelling commitments, encode the background using the shared library, and execute queries to obtain answers. Structural checks cover traceability, satisfiability, and derived-value integrity, while answer execution checks the returned shapes. Figure~\ref{fig:prompt-translation} reproduces excerpts from the task-specific scope instructions and the translation stages of the shared skill. Semantic fidelity checking and expert adjudication are described in Appendix~\ref{sec:audit-records}.

\begin{table}[t]
\centering
\small
\setlength{\tabcolsep}{6pt}
\caption{Expert adjudication outcomes for audited questions, grouped by whether the two encodings initially agreed. Percentages are within each group and rounded independently.}
\label{tab:audit-adjudication}
\begin{tabular}{@{}llr@{}}
\toprule
\textbf{Initial relation} & \textbf{Adjudicated outcome} & \textbf{Within group} \\
\midrule
\multirow{3}{*}{Agree}
 & Both semantically correct & $58.7\%$ \\
 & Both required correction & $27.0\%$ \\
 & Ambiguous source, excluded or clarified & $14.3\%$ \\
\midrule
\multirow{4}{*}{Disagree}
 & One encoding required correction & $81.1\%$ \\
 & Both encodings required correction & $5.3\%$ \\
 & Equivalent answer notation & $9.5\%$ \\
 & Ambiguous source, excluded or clarified & $4.2\%$ \\
\bottomrule
\end{tabular}
\end{table}

\section{Expert Audit}
\label{sec:audit-records}

\subsection{Coverage and fidelity checks}

\paragraph{Coverage.} The census partitions the background into exact character spans. Each substantive claim must be cited by a variable, derived term, or constraint, or explicitly noted with a reason why it does not constrain the answer; headings and purely narrative spans may be marked inert. The check compares \texttt{claim.text} with \texttt{background[start:end]} and reports any uncovered non-whitespace span. It also rejects a model citation to an unknown claim. Coverage therefore checks traceability of the translation, not whether a cited formula is the right one.

\paragraph{Sufficiency (missing rules).} Let $M$ be the assignments admitted by the encoding and $N$ the situations allowed by the Chinese background. A concrete witness in $M\setminus N$ proves under-translation: the model admits a situation forbidden by some sentence. The engine generates admitted worlds, stratified by answer value where feasible, renders each in the background's vocabulary, and asks an independent checker to locate a contradicting source sentence. An abstract instruction to judge the whole model would not identify such a defect.

\paragraph{Necessity (invented rules).} For each named constraint $c$, remove it while leaving the rest of the theory and the relevant question scope fixed. The newly admitted assignments are targeted witnesses in $\mathrm{Mod}(T\setminus\{c\})\setminus M$. If a witness is consistent with the background, $c$ excludes a licensed situation and must be weakened or repaired. If deleting $c$ admits no new world, the audit records whether it still carries load through derived-value coherence or under a counterfactual scope; a baseline no-load finding alone is not a semantic verdict. Derived terms receive separate checks of their chosen output and rejected alternatives because deleting a function definition is not a meaningful ablation.

\subsection{Adjudication of the two encodings}

\paragraph{Expert workflow.} A human logic expert reviews the original item, both encodings, their solver-derived answers, the translation traces, and available check reports. An assisting agent helps locate defects and prepare repairs. Disagreements are traced to the relevant source rules, and agreeing answers are checked for shared misreadings. Repaired encodings are rerun and their answers compared; ambiguous items are clarified or excluded. Only items that pass expert review are included in the dataset.

\paragraph{Audit outcomes.} Table~\ref{tab:audit-adjudication} summarises the outcomes across all audited questions. Among questions with initially agreeing answers, $27.0\%$ required corrections to both encodings, showing that answer agreement can conceal shared misinterpretations. Among questions with initially disagreeing answers, $81.1\%$ required correction to one encoding, and $95.8\%$ reached matching substantive answers after repair or confirmation of equivalent notation.

%% file: sections/10_appendix_evaluation_prompts.tex
\section{Evaluation and Judging Prompts}
\label{sec:evaluation-prompts}

\subsection{Agent Evaluation Prompts}
Figure~\ref{fig:prompt-eval-agent} gives the shared task and the alternative instructions appended for each Hard agent condition. Bracketed placeholders denote item-specific inputs. [SHAPE] specifies the expected JSON answer type: an integer, a Boolean, an array of strings (order does not matter), an array of rows, or an array of \texttt{\{"case": string, "result": value\}} objects.

\subsection{Answer-Judging Prompt}
Figure~\ref{fig:prompt-eval-judge} translates the semantic-equivalence prompt used by DeepSeek V4 Pro for all evaluation conditions. The input includes one entry per sub-question; single-question items use a one-entry list.

%% file: sections/11_appendix_results.tex
\section{Detailed Results Analyses}

\subsection{Base Error Taxonomy}
\label{legacy:sec:base-error-taxonomy}

To characterise model failures on Base, we organise errors into nine categories, whose definitions are provided in Table~\ref{legacy:tab:error-categories}. We then use DeepSeek V4.1 Flash to classify incorrect responses according to this taxonomy, selecting the category that best describes each error. Table~\ref{legacy:tab:base-error-types} presents the resulting distribution of error types for each evaluated model.

\begin{table}[!t]
\centering
\normalsize
\captionsetup{justification=raggedright,singlelinecheck=false}
\setlength{\tabcolsep}{5pt}
\renewcommand{\arraystretch}{1.5}
\caption{Definitions of the nine error categories.}
\label{legacy:tab:error-categories}
\begin{tabularx}{\textwidth}{@{}>{\raggedright\arraybackslash}p{3.2cm}>{\raggedright\arraybackslash}X@{}}
\toprule
\multicolumn{1}{@{}>{\columncolor{casetitlebg}[0pt][\tabcolsep]}l}{\textbf{Category}} & \multicolumn{1}{>{\columncolor{casetitlebg}[\tabcolsep][0pt]}l@{}}{\textbf{Definition}} \\
\midrule
Under-enumeration & Omits valid cases, worlds, branches or members; prunes too early. \\
Over-enumeration & Includes invalid cases; a constraint is left unapplied. \\
Count / aggregation & Finds the right cases but counts, sums, de-duplicates or combines them wrongly. \\
Modal / quantifier & Confuses possible with necessary, some with all, or unique with non-unique. \\
Counterfactual scope & Carries baseline facts into a counterfactual world, or mixes the two. \\
Projection / set level & Confuses elements, sets and families of sets, or the projected fields. \\
Semantic modelling & Misreads a rule, condition, entity, time or evidence status, so the situation is modelled wrongly. \\
Finalisation / format & Reaches the right content but writes the final answer wrongly. \\
Unclear & The material does not separate the causes above. \\
\bottomrule
\end{tabularx}

\par\vspace{20pt}

\centering
\small
\setlength{\tabcolsep}{3pt}
\renewcommand{\arraystretch}{1.4}
\caption{Base error distribution (\%) over three rollouts, classified by DeepSeek V4.1 Flash, within each model's classified errors. Column names follow Table~\ref{legacy:tab:error-categories}.}
\label{legacy:tab:base-error-types}
\begin{tabularx}{\textwidth}{@{}Xrrrrrrrrr@{}}
\toprule
\multicolumn{1}{@{}>{\columncolor{casetitlebg}[0pt][\tabcolsep]}l}{\textbf{Model}} & \cellcolor{casetitlebg}\textbf{Under} & \cellcolor{casetitlebg}\textbf{Over} & \cellcolor{casetitlebg}\textbf{Count} & \cellcolor{casetitlebg}\textbf{Modal} & \cellcolor{casetitlebg}\textbf{CF} & \cellcolor{casetitlebg}\textbf{Proj.} & \cellcolor{casetitlebg}\textbf{Sem.} & \cellcolor{casetitlebg}\textbf{Final.} & \multicolumn{1}{>{\columncolor{casetitlebg}[\tabcolsep][0pt]}l@{}}{\textbf{Unclear}} \\
\midrule
\withlogo{gpt}{GPT-5.6 Sol} & $22.6$ & $27.1$ & $\phantom{0}3.5$ & $\phantom{0}9.5$ & $\phantom{0}4.0$ & $\phantom{0}6.0$ & $20.1$ & $\phantom{0}5.5$ & $\phantom{0}1.5$ \\
\withlogo{claude}{Claude Opus 5} & $41.7$ & $11.8$ & $\phantom{0}4.8$ & $\phantom{0}4.3$ & $\phantom{0}4.3$ & $\phantom{0}2.7$ & $26.2$ & $\phantom{0}3.2$ & $\phantom{0}1.1$ \\
\withlogo{kimi}{Kimi K3} & $49.7$ & $13.8$ & $\phantom{0}4.5$ & $\phantom{0}3.8$ & $\phantom{0}2.0$ & $\phantom{0}3.6$ & $15.4$ & $\phantom{0}5.3$ & $\phantom{0}2.0$ \\
\withlogo{grok}{Grok 4.6} & $44.6$ & $16.2$ & $\phantom{0}7.6$ & $\phantom{0}4.3$ & $\phantom{0}1.1$ & $\phantom{0}4.3$ & $11.9$ & $\phantom{0}5.0$ & $\phantom{0}5.0$ \\
\withlogo{deepseek}{DeepSeek V4 Pro} & $44.4$ & $18.4$ & $\phantom{0}3.6$ & $\phantom{0}4.8$ & $\phantom{0}1.8$ & $\phantom{0}3.8$ & $15.0$ & $\phantom{0}4.2$ & $\phantom{0}4.0$ \\
\withlogo{glm}{GLM-5.3} & $41.4$ & $16.7$ & $\phantom{0}6.6$ & $\phantom{0}4.2$ & $\phantom{0}2.4$ & $\phantom{0}4.7$ & $15.3$ & $\phantom{0}3.1$ & $\phantom{0}5.6$ \\
\withlogo{doubao}{Doubao Seed 2.1 Pro} & $44.0$ & $16.8$ & $\phantom{0}7.1$ & $\phantom{0}3.5$ & $\phantom{0}1.8$ & $\phantom{0}4.1$ & $14.0$ & $\phantom{0}2.6$ & $\phantom{0}6.2$ \\
\bottomrule
\end{tabularx}
\end{table}

\subsection{Accuracy by Hardening Strategy}
\label{sec:strategy-accuracy}

We analyse agent performance across seven models, with six rollouts per model on each question: three in sandbox-only mode and three in library-available mode. A question can involve multiple hardening strategies, so each strategy is evaluated over all sub-questions associated with it. Table~\ref{tab:strategy-accuracy} reports accuracy averaged across models and rollouts separately for the two modes.

\begin{table}[!t]
\centering
\small
\setlength{\tabcolsep}{7pt}
\setlength{\databarwidth}{3.3em}
\setlength{\databarhalfheight}{0.22em}
\renewcommand{\arraystretch}{1.24}
\caption{Agent accuracy (\%) by hardening strategy. Bars share a $0$--$100\%$ scale; bold values mark the highest accuracy in each column.}
\label{tab:strategy-accuracy}
\begin{tabularx}{\textwidth}{@{}Xll@{}}
\toprule
\multicolumn{1}{@{}>{\columncolor{casetitlebg}[0pt][\tabcolsep]}l}{\textbf{Strategy}} & \cellcolor{casetitlebg}\textbf{Sandbox-only} & \multicolumn{1}{>{\columncolor{casetitlebg}[\tabcolsep][0pt]}l@{}}{\textbf{Library-available}} \\
\midrule
Minimal sufficient set & \databar{31.1}{31.1} & \databar{35.1}{35.1} \\
Minimal hitting set & \databar{40.1}{40.1} & \databar{44.9}{44.9} \\
Minimal explanation core & \databar{15.3}{15.3} & \databar{15.3}{15.3} \\
Minimal repair set & \databar{35.1}{35.1} & \databar{37.6}{37.6} \\
Cardinality quota & \databar{35.8}{35.8} & \databar{38.6}{38.6} \\
Statement-truth quota & \databar{35.8}{35.8} & \databar{38.6}{38.6} \\
Conservation balance & \databar{35.8}{35.8} & \databar{38.6}{38.6} \\
Identity-type cycle & \databar{57.1}{\textbf{57.1}} & \databar{52.4}{52.4} \\
Authority-truth cycle & \databar{37.4}{37.4} & \databar{45.9}{45.9} \\
Time-state transition & \databar{32.9}{32.9} & \databar{35.9}{35.9} \\
Role-permission validity & \databar{42.9}{42.9} & \databar{47.3}{47.3} \\
Branching & \databar{35.8}{35.8} & \databar{38.6}{38.6} \\
Distractor premise & \databar{37.5}{37.5} & \databar{40.5}{40.5} \\
Richer answer shape & \databar{35.6}{35.6} & \databar{38.3}{38.3} \\
Uncertainty / multiple answers & \databar{55.6}{55.6} & \databar{58.7}{\textbf{58.7}} \\
\bottomrule
\end{tabularx}
\end{table}

Questions involving minimal explanation cores have the lowest accuracy in both modes, at $15.3\%$, remaining difficult even with access to the logic modeling library. These questions require identifying a jointly inconsistent subset of rules and checking that removing any member restores consistency. The low accuracy suggests difficulty in tracing how constraints jointly produce a contradiction and verifying that every rule in the selected core is necessary. At the easier end, identity-type cycles have the highest sandbox-only accuracy ($57.1\%$), while uncertainty or multiple-answer questions have the highest library-available accuracy ($58.7\%$) and also relatively high sandbox-only accuracy ($55.6\%$). On these evaluated questions, the higher scores suggest relative strengths in reasoning about dependencies between speaker types and statement truth, and in retaining multiple outcomes consistent with the evidence.

\input{sections/12_appendix_limitations}

\input{sections/13_appendix_evaluation_prompt_figures}

\clearpage
\begin{figure}[t]
\centering
\includegraphics[width=\textwidth,height=0.86\textheight,keepaspectratio]{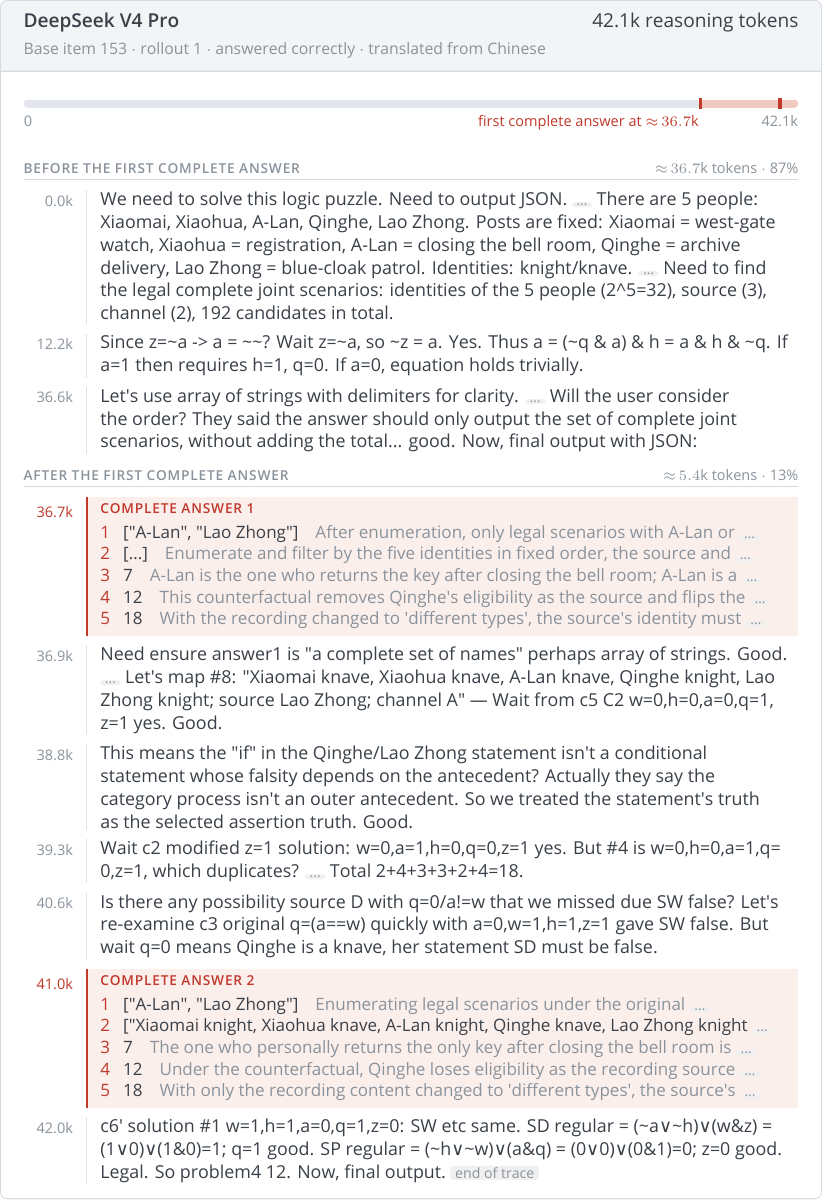}
\caption{A DeepSeek V4 Pro reasoning trace on Base, excerpted and translated from Chinese. The bar shows the whole trace to scale: red ticks mark each complete answer the model writes out, and the shaded part follows the first one. Excerpts keep their original order; the number beside each gives its approximate token position, and grey boxes mark omitted text. Complete answers are shaded, with one line per sub-question: the answer, followed by the start of its reasoning.}
\label{legacy:fig:trace-ds}
\end{figure}

\clearpage
\begin{figure}[t]
\centering
\includegraphics[width=\textwidth,height=0.86\textheight,keepaspectratio]{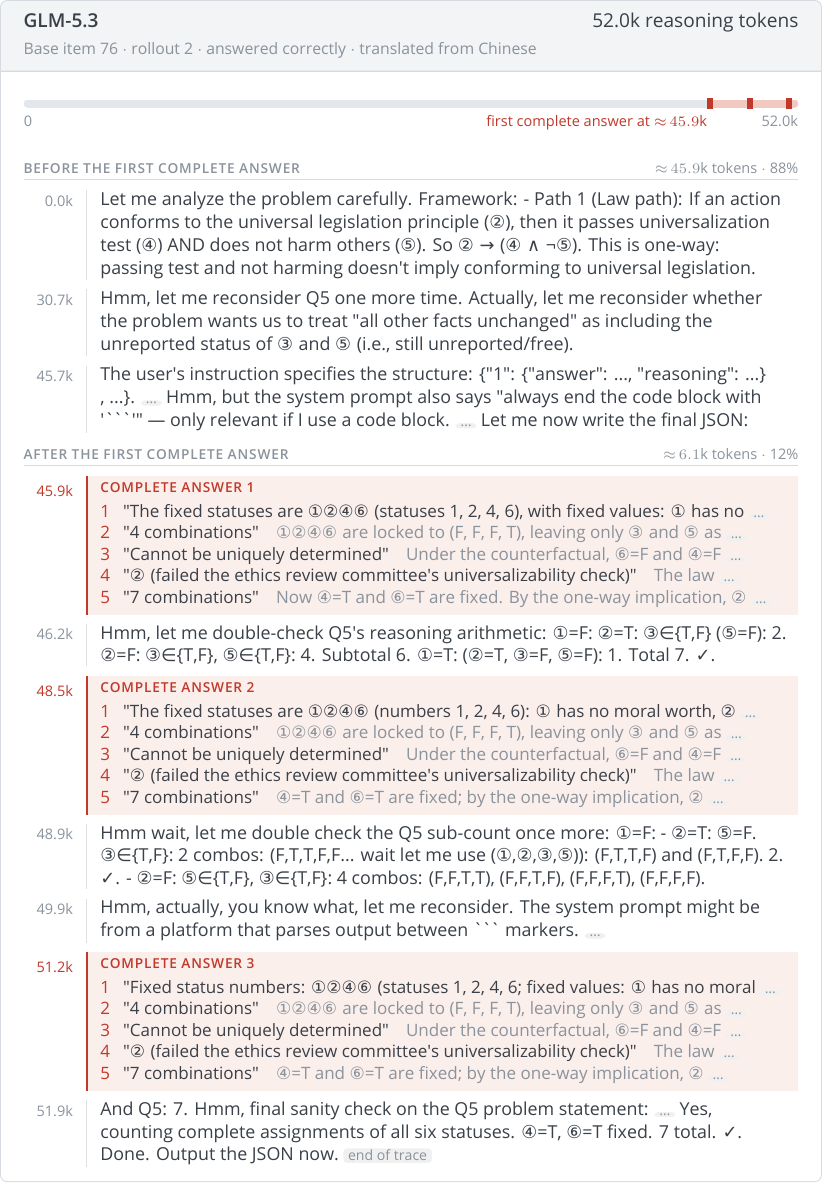}
\caption{A GLM-5.3 reasoning trace on Base, excerpted and translated from Chinese, in the format of Figure~\ref{legacy:fig:trace-ds}: the bar shows the whole trace to scale, numbers beside the excerpts give approximate token positions, grey boxes mark omitted text, and each shaded complete answer shows one line per sub-question.}
\label{legacy:fig:trace-glm}
\end{figure}
\clearpage

\begin{figure}[t]
\centering
\includegraphics[width=\textwidth]{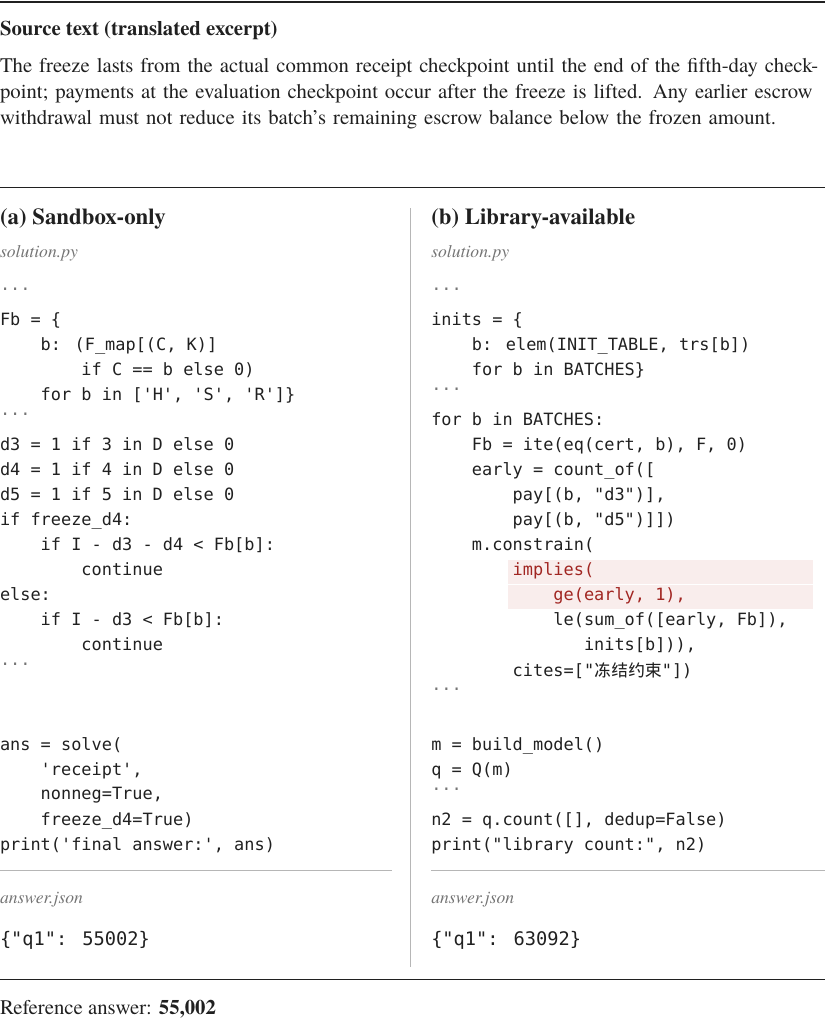}
\caption{Code excerpts and recorded answers from paired DeepSeek V4 Pro runs on Hard (\texttt{r1521\_q1}, rollout 1). The incorrect conditional guard in the library encoding is marked in red. Ellipses mark omitted code; original identifiers are retained, with indentation and line wrapping adjusted for display. The sandbox count matches the reference answer.}
\label{fig:encoding-case}
\end{figure}
\clearpage

%% file: sections/12_appendix_limitations.tex
\section{Limitations}
\label{sec:limitations}

Translating natural-language problems into formal representations remains vulnerable to misinterpretation, and correct solver execution does not guarantee that the resulting answers faithfully reflect the original text. We combine independent translations by two models, multiple layers of auditing, and human-expert review to check semantic fidelity and answer consistency as thoroughly as possible. Nevertheless, these checks cannot provide an absolute guarantee of correctness: shared misinterpretations, overlooked ambiguities, or annotation errors may remain in the released items and reference answers.

%% file: sections/13_appendix_evaluation_prompt_figures.tex
\lstdefinestyle{evaluationprompt}{%
  basicstyle=\fontsize{9}{12}\selectfont\ttfamily,
  frame=single,
  framerule=0.4pt,
  rulecolor=\color{caseframe},
  backgroundcolor=\color{casetitlebg!35},
  breaklines=true,
  breakatwhitespace=true,
  columns=fullflexible,
  keepspaces=true,
  showstringspaces=false,
  numbers=none,
  xleftmargin=0.7em,
  xrightmargin=0.7em,
  framexleftmargin=0.4em,
  framexrightmargin=0.4em,
  aboveskip=2pt,
  belowskip=0pt}

\begingroup
\captionsetup{justification=raggedright,singlelinecheck=false}
\clearpage
\begin{figure}[t]
\centering
\begin{minipage}{\textwidth}
\begin{lstlisting}[style=evaluationprompt,lastline=24]
[Shared task]
# Task

`case/background.txt` states a situation in Chinese and `case/questions.json`
asks one question about it. Answer it.

## What to write

Write your answer to `answer.json` in this directory, an object keyed by
question id:

```json
{"[QID]": <answer>}
```

Answer in Chinese, using the background's own terms for the things it names.
Each question declares the shape its answer takes:

- `[QID]`: [SHAPE]

`answer.json` is the only thing scored.

If you write code, leave your final script as `solution.py`. It is not scored;
it is kept so that how an answer was reached is on the record.
\end{lstlisting}
\par\smallskip
\begin{lstlisting}[style=evaluationprompt,firstline=26,lastline=31]
[Suffix for the sandbox-only agent]
## How you work

You have a shell, a filesystem and network access in this sandbox. Write and run
whatever code helps. Nothing about your approach is prescribed: solve the
questions however you judge best.
\end{lstlisting}
\par\smallskip
\begin{lstlisting}[style=evaluationprompt,firstline=33]
[Suffix for the library-available agent]
## How you work

You have a shell, a filesystem and network access in this sandbox. Write and run
whatever code helps.

A Python library named `translation` is importable here. It is a modelling
language: you declare a problem's unknowns and the constraints among them, and
then ask questions about every assignment that satisfies them, without
enumerating those assignments yourself. `MANUAL.md` in this directory documents
the available declarations and queries.

The library is available for you to use, not required. Nothing about your
approach is prescribed.
\end{lstlisting}
\end{minipage}
\caption{Hard agent task prompt. Section labels in square brackets distinguish the shared task and the two alternative suffixes.}
\label{fig:prompt-eval-agent}
\end{figure}

\clearpage
\begin{figure}[t]
\centering
\begin{minipage}{\textwidth}
\begin{lstlisting}[style=evaluationprompt]
You are a judge of semantic equivalence in Chinese, not a literal string
matcher. Each input item is a complete problem containing multiple
sub-questions. Read the whole problem before deciding, for each sub-question,
whether model_answer is semantically equivalent to reference_answer.
If sets, enumerations, groups, projections, or counts have the same meaning
and differ only in writing order, list order, or set order, mark them correct.
Quantity expressions such as "4", "4 cases", "four cases", and "4 cases in
total" are equivalent; so are answers with the same polarity for yes/no,
can/cannot, and exists/does not exist.
Return false only if model_answer substantively contradicts reference_answer,
omits necessary information, or adds substantively incompatible information.
Return only one JSON object, without Markdown, code blocks, or any additional
explanation. The response must strictly follow this structure:
{"results": [{"id": "bundle_id", "subresults": [
  {"subquestion_no": 1, "match": true, "reason": "brief reason in Chinese"}
]}]}.
match must be a JSON Boolean, true or false, not a string. Return one subresult
for every subquestion in every input item.

The item to judge is below. Return strictly the specified JSON, without any
additional text.
{
  "id": "[ITEM_ID]:[MODEL]",
  "subquestions": [
    {
      "subquestion_no": 1,
      "question": "[QUESTION]",
      "reference_answer": "[REFERENCE ANSWER]",
      "model_answer": "[MODEL ANSWER]"
    }
  ]
}
\end{lstlisting}
\end{minipage}
\caption{Answer-judging prompt and input template (translated).}
\label{fig:prompt-eval-judge}
\end{figure}
\clearpage
\endgroup

%% file: sections/14_appendix_prompt_figures.tex
\lstdefinestyle{hardeningprompt}{%
  basicstyle=\footnotesize\ttfamily,
  frame=single,
  framerule=0.4pt,
  rulecolor=\color{caseframe},
  backgroundcolor=\color{casetitlebg!35},
  breaklines=true,
  breakatwhitespace=true,
  columns=fullflexible,
  keepspaces=true,
  showstringspaces=false,
  numbers=none,
  xleftmargin=0.7em,
  xrightmargin=0.7em,
  framexleftmargin=0.4em,
  framexrightmargin=0.4em,
  aboveskip=2pt,
  belowskip=0pt}

\captionsetup{hypcap=false}
\clearpage
\begin{figure}[t]
\begin{lstlisting}[style=hardeningprompt,firstline=1,lastline=31]
[SYSTEM MESSAGE]

You are the initial_decider for LogicBench reason-mode hardening.

Your job is not to make light suggestions for increasing difficulty. Design an initial hardening blueprint for this item that can directly meet the benchmark hard-case standard.

You must:
- Design from the natural-language background/question/answer in the input only; do not read or cite the formalization.
- Describe planned logical relations in natural language; do not use the symbols U+2192 (implication), U+2227 (and), or U+2228 (or), propositional letters, or formula-like notation.
- Make background_strategy_plans[] and question_strategy_plans[] the principal output.
- Output only a JSON object strictly matching the schema, with no prose, Markdown, explanatory preface, or code fence.
[STRATEGY LIBRARY -- SYSTEM RULES]

[USER MESSAGE]

Strategy-library overview:
[STRATEGY LIBRARY -- CATALOG]

Your task is to design an initial hardening blueprint for this item that directly reaches the benchmark hard-case standard, rather than merely making it a little harder than the source.

Acceptance criteria (the SAME text given to the reviewer and both proposers):

[STRATEGY LIBRARY -- ACCEPTANCE CRITERIA]

Working rules:
- Treat grouped plans as the principal output:
  - background_strategy_plans[]
  - question_strategy_plans[]
- Plan for each qid to correspond, as far as possible, to one independent query. Do not put several duties into one sub-question by default.
- Distinguish a single comparison task that may remain one qid from a bundled multitask request that must be split across qids.
- You are not a proposer. Write a blueprint and duties, not the full background or full question text.
\end{lstlisting}
\caption{Initial decider prompt (translated).}
\label{fig:prompt-decider}
\end{figure}

\clearpage
\begin{figure}[t]
\ContinuedFloat
\begin{lstlisting}[style=hardeningprompt,firstline=32,lastline=52]
[STRATEGY LIBRARY -- WORKING RULES]

The output must satisfy:
- weakness_summary: explain why the original item is not yet a hard case.
- evidence[]: give evidence for that diagnosis, not empty judgments.
- background_strategy_plans[]: each entry has strategy and background_guidance.
- question_strategy_plans[]: each entry has strategy and question_guidance.
- expected_failure_traps[]: list likely shallow paths or mistakes by a model.

background_guidance contains only two fields:
- key_clues: which critical clues to write in the background.
- clue_couplings: [STRATEGY LIBRARY -- FIELD GUIDANCE]

question_guidance contains only one field:
- responsibilities: each entry states the one principal duty of one qid; if several duties are needed, say explicitly that they must be split across qids.

A comparison may be one qid only if the comparison itself is its sole principal task, rather than answering two worlds separately and then supplying a third judgment.
The output must strictly match the response_format schema.

item:
{{item_json}}
\end{lstlisting}
\caption{Initial decider prompt (translated; continued).}
\end{figure}

\clearpage
\begin{figure}[t]
\begin{lstlisting}[style=hardeningprompt]
[SYSTEM MESSAGE]

You are background_proposer_initial for LogicBench hardening.

Your job is to construct the background mechanisms needed for a benchmark hard case, then write them as a complete background and a lightweight dependency table.

You must:
- Write only the background, not answers to the questions.
- Use natural language only in the background. Do not use U+2192, U+2227, U+2228, U+2194, U+00AC, U+22A2, or U+22A8, or encode logical relations with propositional letters or formula-like expressions. Write implication, conjunction, disjunction, equivalence, and negation in words.
- Return the complete background, background_dependency_table, and notes, and output structured JSON only.
[STRATEGY LIBRARY -- SYSTEM RULES]

[USER MESSAGE]

Using the original item and the initial_decider blueprint, generate the first hardening background. Your goal is not to lengthen the background but to construct the hard-case mechanisms first.

Acceptance criteria:
[STRATEGY LIBRARY -- ACCEPTANCE CRITERIA]

Working rules:
- The initial decision contains only grouped plans; compatible strategy_blueprints are generated mechanically by the workflow and are not an additional design source.
- Implement background_strategy_plans[] first, then ensure that those background mechanisms support the high-cost question forms required by question_strategy_plans[].
- Write the background in complete natural-language sentences. Do not use U+2192, U+2227, U+2228, U+2194, U+00AC, U+22A2, or U+22A8, or formula-like abbreviations such as A[U+2192]B or A = true. Write "if ... then ...", "... and ...", "... or ...", "if and only if ...", and "not ..." instead.
- Support questions that map, as far as possible, to one independent query per qid. Do not manufacture difficulty through ambiguous terminology, unclear role boundaries, or omitted rules.
[STRATEGY LIBRARY -- WORKING RULES]

Explicitly forbidden:
[STRATEGY LIBRARY -- RESTRICTIONS]

Output requirements:
- background: the complete background text.
- background_dependency_table: only the most important dependencies, each containing:
  - clue
  - why_it_matters
  - question_hooks
- notes: record the hard-case design focus of this background; avoid empty remarks.
- The output must strictly match the response_format schema.
[STRATEGY LIBRARY -- OUTPUT REQUIREMENTS]

item:
{{item_json}}

initial decision:
{{initial_decision_json}}
\end{lstlisting}
\caption{Initial background proposer prompt (translated).}
\label{fig:prompt-background}
\end{figure}

\clearpage
\begin{figure}[t]
\begin{lstlisting}[style=hardeningprompt,firstline=1,lastline=31]
[SYSTEM MESSAGE]

You are question_proposer_initial for LogicBench hardening.

Your job is to produce a set of benchmark hard-case questions after the background is locked, not merely to add questions that look more complicated.

You must:
- Treat the background text as the facts of the item. The initial plans, dependency table, and notes are non-authoritative design hints only.
- Design each question independently from background + current qid; do not reference or reuse another qid's answer, derived result, complete model table, or private counterfactual. There is no need to copy the shared background into each question.
- Count multitask questions by separately gradable FINAL outputs only. Branch enumeration, counterfactual recomputation, coreference resolution, and evidence checks needed to deliver one final result are internal reasoning.
- Use natural language only in question text. Do not use U+2192, U+2227, U+2228, U+2194, U+00AC, U+22A2, or U+22A8, propositional letters, or formula-like expressions; write every logical relation in complete natural-language sentences.
- Provide minimum evidence of each question's duty through role, background_dependencies, and strategy_tags.
[STRATEGY LIBRARY -- SYSTEM RULES]

[USER MESSAGE]

Using the locked background and initial_decider blueprint, output the first complete question set.

Acceptance criteria:
[STRATEGY LIBRARY -- ACCEPTANCE CRITERIA]

Rules for sources of fact:
- background proposal.background is the only factual authority for the current background.
- background_dependency_table, background notes, and initial plans indicate intended design and possible dependencies; they do not prove that a mechanism, branch, or answer change actually holds.
- When a dependency claim or plan conflicts with the background text, follow the background. Do not design a question on the assumption that a clue has an effect merely because a design claim says so.
- The initial decision contains grouped plans only, without a compatibility mirror or the original item.

Working rules:
- Write text in complete natural-language sentences. Do not use U+2192, U+2227, U+2228, U+2194, U+00AC, U+22A2, or U+22A8, propositional letters, or formula-like abbreviations such as A[U+2192]B or A = true. Express implication, conjunction, disjunction, equivalence, and negation in words.
- Each qid need only be self-contained relative to the shared background: it cannot consume another qid's answer, derived result, or private local counterfactual, and it cannot require first solving another qid to obtain a model table. Shared background and general rules do not need repetition in every question.
- Each qid may demand only one gradable final output and should, where possible, correspond to one independent query. Branch recomputation, coreference resolution, and evidence checks needed for that output are internal reasoning.
\end{lstlisting}
\caption{Initial question proposer prompt (translated).}
\label{fig:prompt-question}
\end{figure}

\clearpage
\begin{figure}[t]
\ContinuedFloat
\begin{lstlisting}[style=hardeningprompt,firstline=32,lastline=70]
- Tolerate isomorphic questions very little:
  - Two questions are duplicates only when their query conditions, target objects, and final-output contracts are substantively the same.
  - Different query operators or projections over one closed state space are valid independent qids; do not call them cosmetic variants merely because they share a model space.
- Do not force several separately deliverable conclusions into one qid, and do not incorrectly split the internal reasoning for a single final output into separate qids.
[STRATEGY LIBRARY -- WORKING RULES]

Explicitly forbidden:
- Explicit multiple sub-questions, such as (1)(2), numbered parts, or "answer these two points separately".
- Implicit multitask questions that demand A, B, and C as separately gradable conclusions.
- A double principal task dressed up as "decide who is authorized + explain whether a variable matters".
- Putting two or more independent conclusions in one qid by writing one long sentence.
- Filler, preparatory, or mere coverage questions.

Rules for comparison questions:
- A single comparison question may remain one qid.
- The comparison itself must be that question's only principal task.
- If the question asks for separate answers for two worlds and then a third judgment, split it into multiple qids by default.

Requirements for writing role:
- Keep it as free text.
- Do not use job titles, generic roles, or empty labels such as Logic Analyst, Reasoner, or Policy Auditor.
- Do not cover two main duties at once. If role needs "both ... and ...", the question probably needs splitting.
[STRATEGY LIBRARY -- ROLE GUIDANCE]

Output requirements:
- Every entry in questions[] must contain:
  - text
  - role
  - background_dependencies
  - strategy_tags
- notes: explain why this question set is at the hard-case level; avoid empty remarks.
- The output must strictly match the response_format schema.
[STRATEGY LIBRARY -- OUTPUT REQUIREMENTS]

initial decision:
{{initial_decision_json}}

background proposal:
{{background_proposal_json}}
\end{lstlisting}
\caption{Initial question proposer prompt (translated; continued).}
\end{figure}

\clearpage
\begin{figure}[t]
\begin{lstlisting}[style=hardeningprompt,firstline=1,lastline=28]
[SYSTEM MESSAGE]

You are the reviewer for LogicBench hardening.

You review the item BEFORE answers are generated; you do not review answers. Your task is not to decide whether the item is merely harder than its source, but whether the candidate has reached the benchmark hard-case standard.

You must:
- Independently audit the final candidate item first and treat it as the only authority for the current item text. The latest revision decision, repair timeline, dependency claims, and question contracts are claims to be checked.
- When auditing a qid, use only the candidate background and that qid. Do not reuse, assume, or obtain a cost reduction from another qid's answer, reasoning, complete model table, or intermediate result.
- Output only proposer-target issues, never answerer-target issues.
- Audit background mechanisms, question-set duties, and grouped-plan coverage in that fixed order.
- Use the fixed issue catalog; do not invent issue codes.
- Apply these fixed boundaries: a real cross-question dependency consumes another qid's derived result; multitask status counts separately gradable final outputs only; leakage means a derived answer or almost complete solution.
- A hard issue must have a reproducible witness. Before raising it, search the candidate for a supposedly missing fact and try to falsify your witness. Do not trigger a hard issue with language such as "possibly", "seemingly", or "please confirm".
- Organize notes into BACKGROUND_AUDIT, QUESTION_AUDIT, and COVERAGE_AUDIT.
[STRATEGY LIBRARY -- SYSTEM RULES]

[USER MESSAGE]

You are performing the answer-blind hard-case gate. Judge the item design only. Do not discuss whether answers are correct, and do not approve merely because the candidate is harder than its source.

[STRATEGY LIBRARY -- ACCEPTANCE CRITERIA]

Rules for sources of fact:
- Independently audit the candidate item first; it is the only authority for the current item text.
- The latest revise decision, repair timeline, background dependency claims, and question contracts are design claims awaiting verification, not proof that a mechanism is present.
- If a candidate domain, mapping, branch, dependency, or difficulty mechanism does not appear explicitly in the candidate or follow from its text, do not approve on the strength of a plan claiming it exists.
- Only after auditing the candidate itself should you compare timeline initial goals, the latest revise decision, and lightweight contracts to see whether the design commitments were realized. If a claim conflicts with the text, follow the candidate and report an issue.
\end{lstlisting}
\caption{Reviewer prompt (translated).}
\label{fig:prompt-reviewer}
\end{figure}

\clearpage
\begin{figure}[t]
\ContinuedFloat
\begin{lstlisting}[style=hardeningprompt,firstline=29,lastline=52]
- In the timeline, decision_plan is intention; actual_question_changes, qid_mapping, and the current candidate show execution. Where a planned qid conflicts with an actual qid, use the actual qid and the current question set. Missing qid numbers, an old index, or a historical planned qid are not natural-language item defects.
- Use the timeline only to verify repairs. When an issue is marked applied and the current candidate has removed it, do not reopen it based on an old plan, qid, index, contract, or strategy tag without a new witness in the current candidate.

Fixed audit order:
1. BACKGROUND_AUDIT
   - Examine the candidate's actual background first. Then verify, one by one, whether this round's background_fixes in the latest revise decision were realized, and whether the current dependency claims match the text.
2. QUESTION_AUDIT
   - Examine each actual question in the candidate first. Then verify, one by one, whether this round's question_fixes in the latest revise decision were realized and whether current question contracts match the question text.
   - Audit each qid independently using candidate background + current qid. A shared background, model space, or general rule is not a cross-question dependency; only a need for another qid's derived result or private local counterfactual counts.
3. COVERAGE_AUDIT
   - Only at this stage, use the timeline's initial background and question goals to check for severe drift in the long-term design. Initial goals are not patches that must each be reimplemented in the current round.
[STRATEGY LIBRARY -- AUDIT CRITERIA]

Every hard issue needs a reproducible witness in its corresponding audit note. For a computable claim, first perform minimal enumeration or a truth-value check and determine whether the candidate already states the allegedly missing fact. Try to falsify the witness. If it fails, reduce the finding to an accurate soft issue or omit it; notes, strengths, and final issues must not contradict one another.

Issue-use rules:
- The reviewer may output ONLY the following 15 proposer issues; internal issues are forbidden.
- If both hard and soft findings have current evidence, issues[] must keep all of them in the same round. Do not omit soft findings merely because a hard finding blocks approval. The workflow uses hard issues to control routing and retries and sends compatible soft issues to the repair node as well.
- Question route (only the question set needs changing, not the background):
  - missing_question_list (hard): The set is empty or has no executable sub-question.
  - cross_question_dependency (hard): A qid must consume another qid's specific derived result or a private local counterfactual that has not been restated. Shared background or model space does not count. The audit note must name the external result consumed and explain why this qid cannot be solved independently.
  - forced_question_type (hard): A forced fit to one of the seven question types creates an unnatural or unsuitable question.
  - multi_task_question (hard): One qid expressly requests two or more separately gradable final outputs. Internal enumeration, recomputation, and verification do not count. The audit note must list those final outputs.
  - question_scope_not_explicit (hard): The background defines a closed candidate domain but the qid fails to specify its query boundary. Give two reasonable interpretations of the query that yield different answers.
\end{lstlisting}
\caption{Reviewer prompt (translated; continued).}
\end{figure}

\clearpage
\begin{figure}[t]
\ContinuedFloat
\begin{lstlisting}[style=hardeningprompt,firstline=53,lastline=80]
  - question_local_rule_rephrasing_overused (soft)
- Decision route (the decider must coordinate repairs to background and questions):
  - missing_background (hard): The background is empty.
  - background_state_space_not_closed (hard): Missing candidate domains, record values, mutual-exclusion rules, or consistency boundaries make enumeration, counting, or projection indeterminate. Give two legal completions that satisfy the written item but yield different answers, or name the missing rule and how it changes the answer.
  - rule_gap_induced_uncertainty (hard): Uncertainty arises from a missing rule, undefined priority, or specification gap. Name the missing rule and its effect on the answer.
  - background_solution_leakage (soft): The background directly states a derived answer, surviving branch, eliminated candidate, target projection or count, or nearly complete solution. Raw mappings and general closure rules do not count.
  - background_bypassed_by_fact_slicing (soft)
  - insufficient_strategy_fusion (soft)
  - insufficient_background_integration (soft)
  - weak_search_space_challenge (soft)
  - difficulty_not_clearly_increased (soft)
- Do not use stem_not_natural_language, unsupported_scope_for_model_enumeration, or local_rule_rephrasing_overused. The latter two were separated by question versus decision root cause.
[STRATEGY LIBRARY -- ISSUE RULES]

Rules for writing issues:
- Each issue must name:
  - A specific witness in the candidate; only quality or coverage findings also name the corresponding strategy or group.
  - The affected qid or sub-question index.
  - Which background clues have failed to matter.
- Also name the clean hard-case standard violated and recommend one of: split, rewrite, or delete.
- A suggestion must prescribe an actionable change, not merely say "increase difficulty".
- A suggestion should normally state the minimum acceptance condition and invariants to preserve. Do not enlarge an unrelated hard issue's minimum repair just to fix a soft issue as well.
- Question-route issues should specify accurate question_ids or subquestion_indexes where possible. A decision-route issue may list affected qids, but it must not disguise a background root cause as a local question defect.
- multi_task_question is a hard proposer issue:
  - If present, send it back to a proposer.
  - Recommend deleting an output or splitting into independent qids only when two final outputs really exist.
  - Do not split internal reasoning needed for one final result into separate questions.
  - It cannot pass as a residual issue.
\end{lstlisting}
\caption{Reviewer prompt (translated; continued).}
\end{figure}

\clearpage
\begin{figure}[t]
\ContinuedFloat
\begin{lstlisting}[style=hardeningprompt,firstline=81,lastline=110]
[STRATEGY LIBRARY -- ISSUE GUIDANCE]

Rules for approved, difficulty_increased, and repair_target:
- approved must not be true if any issue remains.
- difficulty_increased must not be true if the item still fails the benchmark hard-case standard.
- The reviewer reports proposer issues only, so the only permitted repair_target values are:
  - none, when there is no issue.
  - proposer, when there is an issue.

notes must cover exactly three parts:
- BACKGROUND_AUDIT: ...
- QUESTION_AUDIT: ...
- COVERAGE_AUDIT: ...

The output must strictly match the response_format schema.

candidate item:
{{candidate_item_json}}

latest revise decision:
{{revise_decision_json}}

repair timeline:
{{repair_timeline_json}}

background dependency claims:
{{background_dependency_claims_json}}

question contracts:
{{question_contracts_json}}
\end{lstlisting}
\caption{Reviewer prompt (translated; continued).}
\end{figure}
\clearpage

%% file: sections/15_appendix_translation_prompt.tex
\begin{figure}[t]
\begin{lstlisting}[style=hardeningprompt,firstline=1,lastline=43]
[TASK INSTRUCTIONS: EXCERPTS]

# Formalize one item

A Chinese reasoning item is in `case/`. Translate it into a declarative
constraint model, and make that model say what the background says.

[...]

This file is the
assignment and **overrides the skill wherever the two disagree**.

## Scope: Phase 0 through 4 only

Do Phase 0 (census, contract), Phase 1 (commitment sheet), Phase 2 (encode),
Phase 3 (`tl_check`), Phase 4 (`asks.py`, `tl_answer`). Then write your report
and stop.

**Do not run `tl_verify` or `tl_fidelity`, and do not plan around them.** They
are not part of this round. The skill's Phase 5 and Phase 6, and every rejection
code that depends on a checker verdict (`ambiguous_reading`,
`fidelity_not_converged`), do not apply. `checker_unreliable` cannot occur.

[...]

## When are you done

The skill's first discipline says your success signal is the fidelity audit.
**This round has no fidelity audit**, so read that discipline as follows.

[...]

So the condition for stopping is:

1. every claim in `claims.json` is either cited by some constraint or recorded
   in `notes.json` with a category saying why you did not constrain it;
2. every declaration on your commitment sheet is actually true of the code;
3. you have re-read the background one claim at a time against your
   constraints, and `tl_check` is green.

Item 3's first half is the part no tool checks. Do it anyway.

[...]
\end{lstlisting}
\caption{Translation task scope and shared skill (excerpts; omissions marked by [...]).}
\label{fig:prompt-translation}
\end{figure}

\clearpage
\begin{figure}[t]
\ContinuedFloat
\begin{lstlisting}[style=hardeningprompt,firstline=44,lastline=94]

[SHARED SKILL: EXCERPTS]

[...]

### Phase 0: census, lexicon, contract

Split `background.txt` into claims with **character offsets**, so coverage is a
mechanically checkable bijection rather than an impression. Every claim gets an
id. Mark `inert: true` only for sentences with no logical content at all.

[...]

Then declare, in `contract.json`, one `ask` per question with its `modality` and
`returns` shape. Copy the question text **verbatim** from `questions.json`.
Paraphrasing it to silence a modality warning makes the lint check a question
the item does not contain; `tl_check` now errors on that drift.

Modality is one of `enumerate`, `possible`, `necessary`,
`unique`, `count`, `incompatible`. This is a natural-language self-declaration,
not formalization.

[...]

### Phase 1: commitment sheet

Before writing a single constraint, write down, in the experiment document:

- **the background's own roster of what a complete world fixes, if it states
  one.**

[...]

- every domain and why it is closed;
- for each partial function, what happens where it is undefined;
- **for each under-evidenced fact, that it stays a free two-valued variable.**
  Do not add an "unknown" member to any domain. Uncertainty in this corpus is
  epistemic over two-valued facts, carried by leaving a variable unconstrained
  and asking `possible`/`necessary`. An "unknown" value silently deletes real
  models. The only extra members allowed are "not applicable" values, declared
  in `na=[...]`, meaning *no such event*, not *unknown*.

[...]

- for each rule stated as holding "normally", either its declared exception via
  `unless=`, or an explicit statement that you found none;
- for each predicate whose extension is closed by a source-boundary sentence,
  that you relied on the closure;

[...]
\end{lstlisting}
\caption{Translation task scope and shared skill (continued).}
\end{figure}

\clearpage
\begin{figure}[t]
\ContinuedFloat
\begin{lstlisting}[style=hardeningprompt,firstline=95,lastline=143]
### Phase 2: encode

Author `model.py` with `build() -> Model`, importing only from `translation`.
Read [references/primitive_api.md](references/primitive_api.md) first.

Rules the library enforces or the linter checks:

- `constrain()` requires `cites=[claim_id]`. Provenance is mandatory so that a
  missing claim and an invented constraint are both detectable.

[...]

- **One assertion, one `constrain()`.**

[...]

- Give every variable an `alias` in the background's own words, so witnesses
  render in the original vocabulary. A value with no surface form means you
  invented a distinction the background does not make.

[...]

### Phase 3: tl_check, the cheap loop

```bash
python3 code/translation/tl_check.py --case-dir <case_dir>
```

No API calls. Runs lint, solves, checks for degenerate answers, and produces the
ablation load-bearing map. Repair in this order and re-run after each edit:

```text
lint error -> baseline UNSAT -> degenerate answer
```

[...]

`tl_check` therefore evaluates every derive separately with SAT/UNSAT queries.
`derived_undefined`, `derived_domain_unknown`, and
`derived_outside_domain` block the next phase. This check is non-invasive:
an undefined query helper is reported but does not silently delete an otherwise
valid model. Exact per-derive model counts are reading aids and are off by
default. `derived_flags` remains an additional warning for unreachable
`select_unique` labels.

[...]

Stop when `ready_for_fidelity` is true.
\end{lstlisting}
\caption{Translation task scope and shared skill (continued).}
\end{figure}

\clearpage
\begin{figure}[t]
\ContinuedFloat
\begin{lstlisting}[style=hardeningprompt,firstline=144,lastline=192]
### Phase 4: asks.py, the answer

A model that never answers anything is not a translation, and two encodings of
the same item cannot be compared unless both emit an answer in the same shape.
Author `case/asks.py` with one function:

```python
def answers(model):
    """Return {question_id: answer} for every ask in contract.json."""
```

**Every answer must be computed by querying the model you are handed, on this
run.** Not derived by hand, not computed by a separate program and pasted in,
not read off a surrogate model built to stand in for the real one. This is the
most expensive failure the corpus has recorded and no tool catches it: `tl_check`
passes cleanly, because the model is well formed and the query simply never
consults it.

[...]

**Read every status and value back out of `case/answers.json` before you report
it.**

[...]

Rules. The first two are mechanically checked by `tl_answer`; the rest remain
authoring obligations because no deterministic tool can recover query semantics
from Chinese prose:

- One key per `ask` in `contract.json`, no extras, no omissions.
- The value's shape must match that ask's declared `returns`. A `necessary_set`
  ask returns a set of labels; a `count` ask returns an integer; an `enumerate`
  ask returns a list of rows.
- Build the scope with the same `assume` / `world(replace=)` described by the
  verbatim question and the contract note. `tl_answer` cannot infer a semantic
  scope from prose, so `scopes(model)` and the answer program must share one
  helper rather than independently restating the patch.
- Use only the query algebra (`Q(...).necessary_set(...)`, `.count()`,
  `.rows()`, `.possible()`, `.is_unique()`); never post-process with raw Python
  over an enumerated model set. On these items the model set does not fit in
  memory, and the point of the algebra is that it does not have to.

[...]

```bash
python3 code/translation/tl_answer.py --case-dir <case_dir>
```

[...]
\end{lstlisting}
\caption{Translation task scope and shared skill (continued).}
\end{figure}
\clearpage